\pdfoutput=1

\documentclass[11pt]{article}

\usepackage[final]{acl}

\usepackage[utf8]{inputenc}
\usepackage{newtxtext,newtxmath}
\usepackage{microtype}

\usepackage{bm}

\usepackage{graphicx}
\usepackage{booktabs}

\usepackage{tikz}
\usepackage{pgfplots}
\pgfplotsset{compat=1.18}
\usetikzlibrary{arrows.meta,positioning,fit,calc,backgrounds,decorations.pathreplacing}
\usepgfplotslibrary{fillbetween}

\usepackage{latexsym}
\usepackage{pifont}% http://ctan.org/pkg/pifont

\usepackage{linguex}
\usepackage{tcolorbox}

\usepackage{caption}
\usepackage{subcaption}

\newcommand{\pmzerofourspace}{\hphantom{$_{\pm0.00}$}}

\usepackage{placeins}
\usepackage{longtable}

\title{Contrastive Learning with Narrative Twins for Modeling Story Salience}

\author{Igor Sterner, Alex Lascarides
\and Frank Keller \\
  School of Informatics \\
  University of Edinburgh \\
  United Kingdom \\
  {\tt igor.sterner@ed.ac.uk, \{alex,keller\}@inf.ed.ac.uk}
  }

\begin{document}

%%%% spacing for linguex
\setlength{\Exlabelsep}{0.1em}
\setlength{\SubExleftmargin}{1.1em}

\maketitle

\begin{abstract}

Understanding narratives requires identifying which events are most salient for a story's progression. We present a contrastive learning framework for modeling narrative salience that learns story embeddings from \emph{narrative twins}: stories that share the same plot but differ in surface form. Our model is trained to distinguish a story from both its narrative twin and a distractor with similar surface features but different plot. Using the resulting embeddings, we evaluate four narratologically motivated operations for inferring salience (deletion, shifting, disruption, and summarization). Experiments on short narratives from the ROCStories corpus and longer Wikipedia plot summaries show that contrastively learned story embeddings outperform a masked-language-model baseline, and that summarization is the most reliable operation for identifying salient sentences. If narrative twins are not available, random dropout can be used to generate the twins from a single story. Effective distractors can be obtained either by prompting LLMs or, in long-form narratives, by using different parts of the same story.\footnote{\label{fn:data}Code and data are available from \ 
\url{https://github.com/igorsterner/NarrativeTwins}.}

\end{abstract}

\section{Introduction}
\label{sec:intro}

Understanding narratives involves identifying the relative salience of the events in the story (some authors use the equivalent term reportability, \citealt{prince-2003-dictionary, labov-1972-transformation}).
Events that convey information about the protagonists are often more salient than those of minor characters, for example.
The most salient events form a causal chain around which the rest of the story is constructed \citep{labov-1997-further_steps}.

Consider \citeposs{larsson-2005-dragon} murder-mystery novel, \textit{The Girl with the Dragon Tattoo}.
Salander and Blomkvist are solving a decades‑long pattern of murders. 
Example~\ref{ex:dragon-tattoo} is an extract from a summary of a film adaptation.
\citet{papalampidi-2019-movie} annotated the second and third sentences as turning points; in other words, these sentences were determined to be more narratively salient than the surrounding text.

\ex.\label{ex:dragon-tattoo}
\a. Henrik's brother Harald identifies Martin, Harriet's brother and operational head of the Vanger empire, as a possible suspect. 
\b. \textbf{Salander's research uncovers evidence that Martin and his deceased father, Gott\-fried, had committed the murders.}
\c. \textbf{Blomkvist breaks into Martin's house to look for more clues, but Martin catches him and prepares to kill him.} 
\d. Martin brags about having killed women for decades, but denies killing Harriet.

\noindent
NLP applications that process narratives benefit from  modeling salience \citep{zhang-2021-salience}. It has been found to be particularly useful for story summarization \citep{mihalcea-2007-explorations, ladhak-2020-exploring, Papalampidi-2021-movie-summarization, saxena-2024-select}. However, the question of how salience is best operationalized in a computational model has not been explored in a systematic fashion. It is unclear how to compute suitable representations (story embeddings) over which to infer salience, and what type of narrative operation is most suitable for identifying salient sentences.

This paper aims to address these gaps. We propose a model which learns story embeddings by contrasting a narrative with (a)~a narrative which has the same plot but different surface features, and (b)~a distractor which is superficially similar but has a different plot. Using the learned story embeddings, we compare four operations that have been proposed for narrative saliency detection: deletion, shifting, disruption, and summarization. Applying our approach to 
short narratives (ROCStories) and longer Wikipedia plot summaries, we find that story embeddings learned contrastively outperform a masked-language-model baseline, and that summarization is the best operation for saliency detection.

\section{Related Work}
\label{sec:related}

Narratologists generally assume that text is structured around main points that carry most of the meaning, with other points providing extra information to make the text coherent.
In \citeposs{barthes-1966-introduction} theory of narrative structure, he calls the main points cardinal functions and the other points catalyses
(others have used different terminology for similar concepts, e.g., \citealt{mann-1988-rst}).
The cardinal functions are by definition the most narratively salient events, and according to \citet{barthes-1966-introduction} their \textbf{deletion} would change the story's interpretation the most. He also states that these events are the best possible \textbf{summary} of a story.
In contrast, \citet{labov-1967-og} argue that the most narratively salient events are those that are most resistant to \textbf{shifting}, i.e., to occurring anywhere other than their original position in the text. \citet{todorov-1968-grammar} posits that the most salient events express a transition between equilibrium and disequilibrium. We interpret this as a version of surprise \citep{hale-2001-probabilistic}, and call it \textbf{disruption}.

Automatically determining the salience of a textual unit in a larger document is a long-standing research challenge \citep{luhn-1958-automatic, spaerck-1972-tfidf, lin-2000-automated, liu-2018-automatic}. Recent work has trained supervised models based on silver-standard labels derived from units present in a document summary \citep{dunietz-2014-new, liu-2018-automatic, saxena-2024-select}. Alternative approaches use unsupervised learning and model narrative salience via event co‑occurrence around protagonists \citep{chambers-2008-unsupervised} or stylistic change features \citep{ouyang-2015-modeling_mres}. 
\citet{otake-2020-modeling} operationalize \citeposs{barthes-1966-introduction} deletion test with causal language models by measuring the reduction in probability when removing a sentence.
Building on this, \citet{wilmot-2021-salience} apply the deletion test to book-length narratives based on a model that combines bi-directional story embeddings with retrieval-augmented generation and a memory model. They compare deletion with shifting, and find that shifting performs worse.
Using bi-directional representations is also intuitive, because the salience of an event in a narrative may only become apparent with knowledge of the future context.
Global similarity approaches rank sentences by centrality \citep[e.g., TextRank,][]{mihalcea-2004-textrank} or by similarity to a whole-document representation \citep{zhong-2020-extractive}, and can be seen as an instantiation of \citeposs{barthes-1966-introduction} claim that the most narratively salient events will be the best summary of a story.

\section{Contrastive Learning for Narratives}
\label{sec:embeddings}

\begin{figure*}
\centering
\begin{subfigure}[b]{0.45\textwidth}
\centering
\begin{small}
\begin{tikzpicture}[
  >={Latex[length=1.5mm]},
  line width=.25pt,
  block/.style={draw, rounded corners, minimum width=1cm, minimum height=1cm, align=center},
  colbox/.style={draw, rounded corners, inner xsep=8pt, inner ysep=9pt},
]

\def\yt{1.0}
\def\yb{-1.0}

\node[align=center] (x) at (0,0) {underlying\\story};

\node[block] (v)  at (2,\yt) {$\mathbf{x}$};
\node[block] (yT) at (4,\yt) {$\mathbf{y}$};

\node[block] (vp) at (2,\yb) {$\mathbf{x^+}$};
\node[block] (yX) at (4,\yb) {$\mathbf{y^+}$};

\node[colbox, fit=(v) (vp), label={[align=center,yshift=1.5mm]north:{narrative\\twin}}] (colV) {};
\node[colbox, fit=(yT) (yX), label={[align=center,yshift=1.5mm]north:narrative\\embedding}] (colY) {};

\node[fit=(v)(yT), inner xsep=8pt, inner ysep=9pt] (topband) {};
\node[fit=(vp)(yX), inner xsep=8pt, inner ysep=9pt] (botband) {};

\begin{scope}[on background layer]
  \fill[gray!20, rounded corners] (topband.north west) rectangle (topband.south east);
  \fill[gray!20,  rounded corners] (botband.north west) rectangle (botband.south east);
\end{scope}

\draw[->] (x.east) -- (v.west);
\draw[->] (x.east) -- (vp.west);

\draw[->] (v) -- node[midway,above] {$f_\theta$} (yT);
\draw[->] (vp) -- node[midway,above] {$f_\theta$} (yX);

\coordinate (topE) at (5.5,\yt);
\coordinate (botE) at (5.5,\yb);

\draw[->] (yT.east) -- (topE);
\draw[->] (yX.east) --(botE);

\draw[-, dashed] (topE) .. controls +(1.2,0) and +(1.2,0) .. (botE);

\def\ydist{-3.3}

\node[align=center] (xd) at (0,\ydist) {different\\underlying\\story};

\node[block] (vd)  at (2,\ydist) {$\mathbf{x}^-$};
\node[block] (yd)  at (4,\ydist) {$\mathbf{y}^-$};

\node[fit=(vd)(yd), inner xsep=8pt, inner ysep=9pt] (distband) {};
\begin{scope}[on background layer]
  \fill[gray!20, rounded corners] (distband.north west) rectangle (distband.south east);
\end{scope}

\node[colbox, fit=(vd), label={[align=center,yshift=1.5mm]north:{distractor}}] {};
\node[colbox, fit=(yd), label={[align=center,yshift=1.5mm]north:{\ \\ \ }}] {};

\draw[->] (xd.east) -- (vd.west);
\draw[->] (vd) -- node[midway,above] {$f_\theta$} (yd);

\coordinate (botd) at (5.5,\ydist);

\draw[->] (yd.east) --(botd);
\draw[-, dashed] (botd) .. controls +(1.5,0) and +(1.5,0) .. (topE);

\end{tikzpicture}
\end{small}
\caption{Learning representations of narratives}
\label{fig:model}
\label{fig:training}
\end{subfigure}
\hspace{0.05\textwidth}
\begin{subfigure}[b]{0.45\textwidth}
\centering
\begin{small}
\begin{tikzpicture}[
  >={Latex[length=1.5mm]},
  line width=.25pt,
  block/.style={draw, rounded corners, minimum width=0.95cm, minimum height=0.95cm, align=center, outer sep=0pt},
  labeltiny/.style={font=\scriptsize, inner sep=1pt},
  labeltinyit/.style={font=\scriptsize\itshape, inner sep=1pt}
]

\def\W{8.0}
\def\H{5.0}

\def\Lxy{0.5}
\def\Lout{0.4}

\draw (\W/2+1.5,1) -- (\W/2+1.5,\H+1.5);
\draw (1.3,\H/2+1.5) -- (\W+1,\H/2+1.5);

\def\xL{1.8}
\def\xLr{6.2}

\def\yTLtop{5.8}
\def\yTLbot{4.7}
\def\yTRtop{5.8}
\def\yTRbot{4.7}

\def\yBLtop{2.6}
\def\yBLbot{1.4}
\def\yBRtop{2.6}
\def\yBRbot{1.4}

\node[block] (x_del_top) at (\xL,\yTLtop) {$\mathbf{x}_{1:N}$};
\node[block, anchor=west] (y_del_top) at ($(x_del_top.east)+(\Lxy,0)$) {$\mathbf{y}$};
\path (x_del_top.east) -- (y_del_top.west) coordinate[pos=.5] (delTopMid);
\draw[-] (x_del_top.east) -- (delTopMid);
\draw[->] (delTopMid) -- (y_del_top.west);
\node[above,yshift=1pt] at (delTopMid) {$f_\theta$};
\coordinate (delTopE) at ($(y_del_top.east)+(\Lout,0)$);
\draw[->] (y_del_top.east) -- (delTopE);

\node[block] (x_del_bot) at (\xL,\yTLbot) {$\mathbf{x}_{\substack{1:N \\ \setminus j:k}}
$};
\node[block, anchor=west] (y_del_bot) at ($(x_del_bot.east)+(\Lxy,0)$) {$\mathbf{y}^\prime$};
\path (x_del_bot.east) -- (y_del_bot.west) coordinate[pos=.5] (delBotMid);
\draw[-] (x_del_bot.east) -- (delBotMid);
\draw[->] (delBotMid) -- (y_del_bot.west);
\node[above,yshift=1pt] at (delBotMid) {$f_\theta$};
\coordinate (delBotE) at ($(y_del_bot.east)+(\Lout,0)$);
\draw[->] (y_del_bot.east) -- (delBotE);

\draw[-, dashed] (delTopE) .. controls +(0.9,0) and +(0.9,0) .. (delBotE);

\node[block] (x_sum_top) at (\xLr,\yTRtop) {$\mathbf{x}_{1:N}$};
\node[block, anchor=west] (y_sum_top) at ($(x_sum_top.east)+(\Lxy,0)$) {$\mathbf{y}$};
\path (x_sum_top.east) -- (y_sum_top.west) coordinate[pos=.5] (sumTopMid);
\draw[-] (x_sum_top.east) -- (sumTopMid);
\draw[->] (sumTopMid) -- (y_sum_top.west);
\node[above,yshift=1pt] at (sumTopMid) {$f_\theta$};
\coordinate (sumTopE) at ($(y_sum_top.east)+(\Lout,0)$);
\draw[->] (y_sum_top.east) -- (sumTopE);

\node[block] (x_sum_bot) at (\xLr,\yTRbot) {$ \mathbf{x}_{j:k}$};
\node[block, anchor=west] (y_sum_bot) at ($(x_sum_bot.east)+(\Lxy,0)$) {$\mathbf{y}^\prime$};
\path (x_sum_bot.east) -- (y_sum_bot.west) coordinate[pos=.5] (sumBotMid);
\draw[-] (x_sum_bot.east) -- (sumBotMid);
\draw[->] (sumBotMid) -- (y_sum_bot.west);
\node[above,yshift=1pt] at (sumBotMid) {$f_\theta$};
\coordinate (sumBotE) at ($(y_sum_bot.east)+(\Lout,0)$);
\draw[->] (y_sum_bot.east) -- (sumBotE);

\draw[-, dashed] (sumTopE) .. controls +(0.9,0) and +(0.9,0) .. (sumBotE);

\node[block] (x_shi_top) at (\xL,\yBLtop) {$\mathbf{x}_{1:N}$};
\node[block, anchor=west] (y_shi_top) at ($(x_shi_top.east)+(\Lxy,0)$) {$\mathbf{y}$};
\path (x_shi_top.east) -- (y_shi_top.west) coordinate[pos=.5] (shiTopMid);
\draw[->] (shiTopMid) -- (y_shi_top.west);
\draw[-] (x_shi_top.east) -- (shiTopMid);
\node[above,yshift=1pt] at (shiTopMid) {$f_\theta$};
\coordinate (shiTopE) at ($(y_shi_top.east)+(\Lout,0)$);
\draw[->] (y_shi_top.east) -- (shiTopE);

\node[block] (x_shi_a) at (\xL,\yBLbot+0.15) {$\mathbf{x}_{1:N}^{i,0}$};
\node[block, opacity=0.45] (x_shi_b) at (\xL+0.65,\yBLbot-0.55) {$\mathbf{x}_{1:N}^{i,t}$};

\node[block, anchor=west] (y_shi_a) at ($(x_shi_a.east)+(\Lxy,0)$) {$\mathbf{y}^\prime_0$};
\node[block, anchor=west, opacity=0.45] (y_shi_b) at ($(x_shi_b.east)+(\Lxy,0)$) {$\mathbf{y}^\prime_t$};

\path (x_shi_a.east) -- (y_shi_a.west) coordinate[pos=.5] (shiAMid);
\draw[-] (x_shi_a.east) -- (shiAMid);
\draw[->] (shiAMid) -- (y_shi_a.west);
\node[above,yshift=1pt] at (shiAMid) {$f_\theta$};

\path (x_shi_b.east) -- (y_shi_b.west) coordinate[pos=.5] (shiBMid);
\draw[-, opacity=0.8] (x_shi_b.east) -- (shiBMid);
\draw[->, opacity=0.8] (shiBMid) -- (y_shi_b.west);
\node[above,yshift=1pt,opacity=0.9] at (shiBMid) {$f_\theta$};

\coordinate (shiAE) at ($(y_shi_a.east)+(\Lout,0)$);
\coordinate (shiBE) at ($(y_shi_b.east)+(\Lout,0)$);
\draw[->] (y_shi_a.east) -- (shiAE);
\draw[->, opacity=0.45] (y_shi_b.east) -- (shiBE);

\draw[-, dashed]
  let \p1=(shiTopE), \p2=(shiAE),
      \n1={veclen(\x1-\x2,\y1-\y2)/2},
      \n2={atan2(\y1-\y2,\x1-\x2)}
  in
  (shiTopE) arc[start angle=\n2, delta angle=-180, radius=\n1];

\draw[-, dashed, black!50]
  let \p1=(shiTopE), \p2=(shiBE),
      \n1={veclen(\x1-\x2,\y1-\y2)/2},
      \n2={atan2(\y1-\y2,\x1-\x2)}
  in
  (shiTopE) arc[start angle=\n2, delta angle=-180, radius=\n1];

\node[block] (x_sur_pre) at (\xLr,\yBRtop) {$\mathbf{x}_{1:k}$};
\node[block, anchor=west] (y_sur_pre) at ($(x_sur_pre.east)+(\Lxy,0)$) {$\mathbf{y}$};
\path (x_sur_pre.east) -- (y_sur_pre.west) coordinate[pos=.5] (surPreMid);
\draw[-] (x_sur_pre.east) -- (surPreMid);
\draw[->] (surPreMid) -- (y_sur_pre.west);
\node[above,yshift=1pt] at (surPreMid) {$f_\theta$};
\coordinate (surTopE) at ($(y_sur_pre.east)+(\Lout,0)$);
\draw[->] (y_sur_pre.east) -- (surTopE);

\node[block] (x_sur_post) at (\xLr,\yBRbot) {$\mathbf{x}_{1:j-1}$};
\node[block, anchor=west] (y_sur_post) at ($(x_sur_post.east)+(\Lxy,0)$) {$\mathbf{y}^\prime$};
\path (x_sur_post.east) -- (y_sur_post.west) coordinate[pos=.5] (surPostMid);
\draw[->] (surPostMid) -- (y_sur_post.west);
\draw[-] (x_sur_post.east) -- (surPostMid);
\node[above,yshift=1pt] at (surPostMid) {$f_\theta$};
\coordinate (surBotE) at ($(y_sur_post.east)+(\Lout,0)$);
\draw[->] (y_sur_post.east) -- (surBotE);

\draw[-, dashed] (surTopE) arc[start angle=90, end angle=-90, radius={(\yBRtop-\yBRbot)/2}];

\node[font=\small] at (\xL,\H+1.7) {deletion};
\node[font=\small] at (\xLr+0.4,\H+1.7) {summarization};
\node[font=\small] at (\xL,\H/2+1.0) {shifting};
\node[font=\small] at (\xLr+0.1,\H/2+1.0) {disruption};

\end{tikzpicture}
\end{small}
\vspace{-0.3cm}
\caption{Four modes of inferring salience (see Eqs. \ref{equ:del}--\ref{equ:sum})}
\label{fig:inference}
\end{subfigure}
\caption{Left: Overview of contrastive training for learning narrative embeddings. Inputs are an anchor narrative $\mathbf{x}$, its narrative twin $\mathbf{x^+}$, and a distractor $\mathbf{x^-}$. A shared encoder $f_\theta$ generates embeddings. The embeddings are optimized with a contrastive loss to be similar for twins and dissimilar to distractors (and in-batch negatives). Right: At inference, we score the salience of each sentence via four theoretically motivated operations.}
\end{figure*}
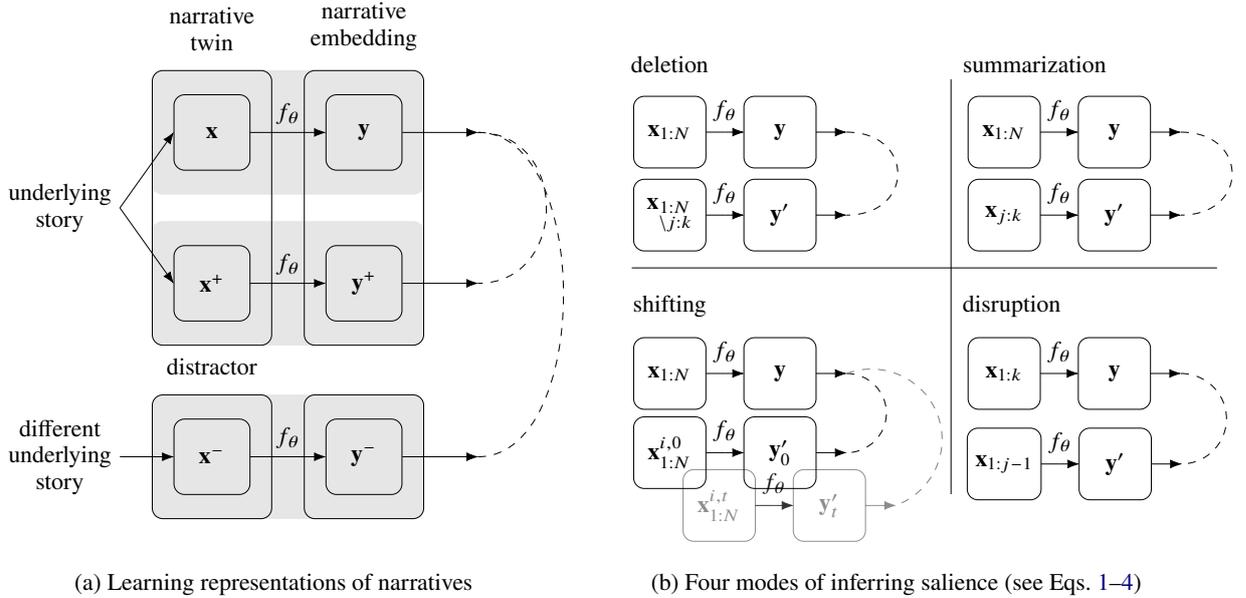

Text representations are standardly learned using a masked or causal language modeling objective. The resulting model can then be fine-tuned for a particular task using contrastive learning. In this work, we will test the hypothesis that better story representations can be obtained by training a model contrastively to distinguish between an original narrative, a narrative twin, and a distractor.
Narrative twins are two versions of the same story.
These two versions could be two translations of a story, a story and an adaptation of it, two summaries of the same story, and so on.
The crucial ingredient is that the versions share the same plot but differ in surface form.
The distractors are narratives that have similar surface features but a different plot.

\citet{hatzel-2024-tell} provide narrative twins in their \textit{Tell me again!} dataset. It consists of plot summaries from different language versions of Wikipedia. The narratives from non-English versions were machine-translated into English and back-translations are filtered out \citep[see Figure 2 in][for example narrative twins]{hatzel-2024-tell}. \citet{hatzel-2024-story} fine-tune a language model with a contrastive objective, using narrative twins as positives and in-batch narratives as negatives. A problem that arises is that the narrative twins have large and shallow lexical overlap, e.g., in proper nouns. To counter this, \citet{hatzel-2024-story} propose an augmentation strategy that substitutes entities in the narrative twins with random alternatives. They report on the usefulness of the resulting learned embeddings on various retrieval tasks. In our work, we will build on their overall approach, but apply it to determining salient sentences in narratives, and in particular, we will compare the performance of theoretically motivated salience operations (deletion, shifting, summarization, disruption) on contrastively learned narrative representations.

We will also pursue an alternative approach in which narrative twins are not based on different textual versions of the same story, but on different representations of a single, textually identical story. 
We use an approach that utilizes random dropout \citep{gao-2021-sim_cse, liu-2021-fast}, and we call the result \emph{dropout twins}.
Dropout twins are two similar embeddings for the same story, generated by passing the story through a neural model twice with different random dropout. 
We hypothesize that dropout twins approximate narrative twins at the representation level.

In contrastive learning, the negative pairs (the distractors) also play an important role \citep{xu-2022-negatives}.
\citet{hatzel-2024-story} use random in-batch negatives, i.e., unrelated narratives.
Their assumption is thus that distinguishing a story's narrative twin from many unrelated stories requires a complete understanding of the story.
\citet{mu-2024-generating} argue that this assumption is implausible, and that random negative narratives are too easily distinguishable.
For their task of modeling narrative coherence, they show that hard negative samples with lower coherence can be generated by applying perturbations to positive samples.
For our task of modeling narrative salience, we hypothesize that the best negative samples will be coherent narratives with similar surface features but a different plot.

\section{The Model}
\label{sec:model}

Our experiments address two research questions: (1)~Which of the theoretically proposed operations (deletion, shifting, disruption, summarization) best predicts narrative salience? (2)~Does a model trained to make embeddings of narrative twins similar perform better under these operations than a model that does not exploit narrative twins? 

Figure~\ref{fig:model} shows the model that we propose for the task.
The inputs to the model are an anchor narrative $\mathbf{x}$, its narrative twin $\mathbf{x^+}$, and a distractor negative $\mathbf{x^-}$.
The key requirement for the narrative twin is that it has the same underlying story as the anchor narrative.
In this respect, the twin could also be a dropout twin; this removes the burden of narrative twins needing to be available for training.
Meanwhile, the key requirement for the distractor negative is that superficially it is similar to the anchor, i.e., the same characters and setting, similar actions and events.
However, in the distractor the underlying story must be different.

For each narrative, $\mathbf{x}$, $\mathbf{x^+}$ and $\mathbf{x^-}$ is transformed independently with a bi-directional embedding model $f_\theta$ into story embeddings $\mathbf{y}$, $\mathbf{y^+}$ and $\mathbf{y^-}$.
The learning objective is a contrastive loss that makes $\mathbf{y}$ similar to $\mathbf{y^+}$ and different from $\mathbf{y^-}$ as well as from all other in-batch stories.
In all of our experiments we use the standard InfoNCE loss \citep{gutmann-2010-noise,oord-2019-info_nce}.

We use the resulting story embeddings to infer narrative salience and operationalize the four narrative operations as follows (see Figure~\ref{fig:inference}).
Consider a narrative consisting of a sequence of tokens, $\mathbf{x}_{1:n}$.
Let $s_i$ be the salience of sentence $i$, which is tokenized into indices $j$ until $k$ (inclusive) in $\mathbf{x}$.

\noindent\textbf{Deletion:} similarity between the embedding of the whole story and embedding of the story with one sentence removed:
\begin{equation}\label{equ:del}
    s_i^{del} = 1-\operatorname{sim}\left( f_\theta (\mathbf{x}_{1:N}), f_\theta(\mathbf{x}_{1:N \setminus j:k})\right)
\end{equation}
\noindent\textbf{Shifting:} average similarity between the embedding of the whole story and the embedding of that story with sentence $i$ shifted to each of every other position. Let $\mathbf{x}^{i,t}$ be $\mathbf{x}$ with sentence $i$ shifted to be prior to sentence $t$:
\begin{equation} 
s_i^{shi} = 1-\frac{1}{Z_i} \sum_{t \neq i} \operatorname{sim}\left( f_\theta( \mathbf{x}_{1:N} ), f_\theta( \mathbf{x}_{1:N}^{i,t}) \right)
\end{equation}
\noindent\textbf{Disruption:} similarity between the story up to and including sentence $i$, and up to but not including sentence $i$ (recall $i$ is tokenized into indices $j$ to~$k$):\footnote{This implementation offers a measure specific to the contribution of sentence $i$ and is a standard formulation of surprisal extended to the sentence-level.}
\begin{equation}\label{equ:sur}
    s_i^{dis} = 1-\operatorname{sim}\left( f_\theta( \mathbf{x}_{1:k} ), f_\theta( \mathbf{x}_{1:j-1} ) \right)
\end{equation}
\noindent\textbf{Summarization:} similarity between the embedding of the whole story and the embedding of sentence~$i$:
\begin{equation}\label{equ:sum}
    s_i^{sum} = \operatorname{sim}\left( f_\theta( \mathbf{x}_{1:N} ), f_\theta( \mathbf{x}_{j:k} ) \right)
\end{equation}
A range of similarity measures can be applied in these operationalizations of salience; here, we use cosine similarity, which is widely used for comparing embeddings.

Our training and inference methods need to be extended for long narratives, for two reasons: 
(1)~As the narratives get longer, the resulting embeddings are less likely to represent the whole input narrative (as opposed to a subsection of it).
(2)~The operations used for inference of narrative salience scale in $O(N)$ (deletion and disruption) or $O(N^2)$ (shifting) time with the number of input sentences~$N$.

We address this by adjusting the resolution at which the contrastive comparisons are being made.
For intuition, consider learning from a whole book, a narrative twin of the book, and a negative distractor.
Rather than comparing embeddings of the whole books, our proposed intervention is to compare embeddings at, for instance, the chapter level. 
The role of $f_\theta$ is thus to generate contextualized chapter embeddings, and Chapter $c$ of the book is compared with Chapter $c$ of the twin and Chapter $c$ of the negative distractor.
In the following, we will refer to \emph{windows} instead of chapters.

Our approach is operationalizable with current embedding models, because they generate a contextualized embedding for each token in the input.
$f_\theta$ thus pools token representations for each window \citep[see][]{gunther-2025-latechunking} in aligned $\mathbf{y}$, $\mathbf{y^+}$ and $\mathbf{y^-}$.
The contrastive comparison for training is then made at the window level.
The operations used for inference are also applied at the window level, e.g., for deletion we delete each sentence in the window one at a time and compare window embeddings before and after the operation.

Different windows within a single narrative involve the same characters, will frequently have similar settings, and yet convey a different plot.
We hypothesize that the different windows within a narrative can serve as the distractors in the contrastive comparisons \citep[see][]{conti-2025-context}.
We will call them \emph{in-story negatives}.

\section{Experiment 1: ROCStories}
\label{sec:roc-model}

Our first experiment is as simple and controlled as possible, using short-form narratives.
We use the ROCStories dataset \citep{mostafazadeh-2016-roc_stories} of human-written, English five-sentence stories, such~as:

\ex.\label{ex:roc-example}
\a. Ricky was playing tennis with Bob. \label{ex:ricky-was-1}
\b. \textbf{Bob served the ball and hit Ricky right in the head!} \label{ex:ricky-was-2}
\c. Ricky fainted on the tennis court.
\d. Bob gave Ricky some water to drink.
\e. Ricky felt better after resting for awhile.

Sentence~\ref{ex:ricky-was-2} is the narrative peak of the story; it describes the most important, most salient event.
Removing this sentence (deletion) affects our interpretation of the story far more than removing other sentences, because we would be left confused as to why the subsequent events occurred.
The sentence also summarizes the entire story well (summarization), is surprising (disruption) and -- perhaps more debatably -- is most resistant to being shifted to any other position in the story (shifting).

In many stories, the presence of a single narrative peak sentence is less clear than in the above example, because the peak spans more than one sentence or there is no unique clear peak.
Our approach follows prior work on narrative salience \citep{liu-2018-automatic,otake-2020-modeling,wilmot-2021-salience} in viewing the problem as a ranking task.

\subsection{Annotation Methodology}
\label{sec:annotation}

To evaluate our models, we collect human annotation of what participants determine is the most salient sentence in each five-sentence story. This is an easier task than asking annotators to produce a full ranking of all sentences.
For each story, we collect multiple such annotations, and then compute a ranking of each sentence based on how many annotators selected it. 
The full guidelines given to participants can be found in Appendix~\ref{sec:guidelines}.

Each of 500 stories received five annotations from crowd-workers (10 annotations were obtained from each of 250 US/UK crowd-workers).
All participants are required to complete a comprehension check prior to the study in which they match five summaries with five stories.
They then each annotate 10 stories.
We only keep data from participants who passed the comprehension test ($236/250 = 94.4\%$ of participants).
We split the evaluation set in half for validation and testing.
The dataset along with the human-written summaries is publicly available (see Footnote~\ref{fn:data}), but we will not use the summaries further here.

The mean entropy over the labels in each story is 1.02, corresponding to a perplexity of $2^{1.02}\approx2.03$.
This shows that on average annotators behave as if there are approximately two equally likely options per story.
For a measurement of human agreement on the task, a subset of 50 of the test stories was double annotated.
This involved collecting a further five annotations for each story in the subset (a further 250 annotations). 
Agreement was measured at $\rho=0.50$ (Spearman rank correlation), indicating reasonable agreement.
Evidently there is some subjectivity in the task, and we assume that limiting the human annotation to five ratings per story introduces some noise.

\subsection{Metrics and Significance Testing}
\label{sec:metrics}

As our metric of performance, we compare the ranking from a model's salience predictions with the human ranking, again using Spearman's rank correlation.
As a secondary measure, we compute area under the curve (AUC).
AUC is a metric from information retrieval which measures the proportion of relevant and irrelevant sentences for which the predicted saliency of the relevant sentence is greater than that of the irrelevant sentence.
For our purposes, we consider a sentence with at least one human saliency label as relevant; all others are irrelevant.

In this paper, we test differences in performance for significance with a paired, two-tailed permutation test, approximated by $N=10{,}000$ with significance threshold $\alpha=0.05$.
We repeat all training runs across three random seeds, and repeat LLM prompts three times; the significance tests are based on the median runs.
In our tables, systems that are statistically better than the most other systems within a group are in bold.

\subsection{Narrative Twins and Negatives}
\label{sec:narrative_twins}

When the authors of the ROCStories wrote five-sentence stories, they most likely imagined a plot that is more elaborate than the final events they expressed in the story.
With this inspiration, we first prompt an LLM (\textsc{GPT-4.1 mini}) to generate a verbose version of the story, in which many more possible events are verbalized.
We then independently prompt the same LLM to generate a five-sentence story based on the verbose story, which serves as the narrative twin.
For the negatives, we prompt the same LLM to generate a new five-sentence story that changes the main event and keeps other events similar.
We generate twins and negatives for $20{,}000$ ROCStories, resulting in our training corpus.
Prompts and example outputs are in Appendix~\ref{sec:rocstory:prompts}.

\begin{table*}
\centering
\caption{Results on ROCStories test set. We report mean $\pm$ one standard deviation (computed across three different random seeds). In bold are the winning results according to permutation-based significance tests (see main text).}
\label{tab:roc-main}
\begin{small}
\begin{tabular}{lrrrrrrrr}
\toprule
& \multicolumn{2}{c}{\textbf{Deletion}} & \multicolumn{2}{c}{\textbf{Shifting}} & \multicolumn{2}{c}{\textbf{Disruption}} & \multicolumn{2}{c}{\textbf{Summarization}} \\
\cmidrule(lr){2-3} \cmidrule(lr){4-5} \cmidrule(lr){6-7} \cmidrule(lr){8-9} 
& \multicolumn{1}{c}{\textbf{$\bm{\rho}$}} & \multicolumn{1}{c}{\textbf{AUC}} & \multicolumn{1}{c}{\textbf{$\bm{\rho}$}} & \multicolumn{1}{c}{\textbf{AUC}} & \multicolumn{1}{c}{\textbf{$\bm{\rho}$}} & \multicolumn{1}{c}{\textbf{AUC}} & \multicolumn{1}{c}{\textbf{$\bm{\rho}$}} & \multicolumn{1}{c}{\textbf{AUC}} \\ \midrule
\textsc{Masked LM} & 0.20$_{\pm0.01}$ & 0.61$_{\pm0.01}$ & 0.17$_{\pm0.04}$ & 0.59$_{\pm0.02}$ & -0.14$_{\pm0.00}$ & 0.42$_{\pm0.00}$ & 0.21$_{\pm0.01}$ & 0.62$_{\pm0.01}$ \\
\textsc{Dropout Twins} & 0.37$_{\pm0.01}$ & 0.70$_{\pm0.01}$ & 0.25$_{\pm0.01}$ & 0.63$_{\pm0.00}$ & 0.06$_{\pm0.01}$ & 0.53$_{\pm0.01}$ & 0.35$_{\pm0.01}$ & 0.70$_{\pm0.00}$ \\
\textsc{Narrative Twins} & 0.38$_{\pm0.00}$ & 0.70$_{\pm0.00}$ & 0.31$_{\pm0.01}$ & 0.66$_{\pm0.00}$ & 0.13$_{\pm0.00}$ & 0.57$_{\pm0.00}$ & \textbf{0.44$_{\pm0.01}$} & \textbf{0.76$_{\pm0.00}$} \\
\bottomrule
\end{tabular}
\end{small}

\end{table*}

\begin{table}[h]
\caption{ROCStories human, baseline and LLM prompting results on the test set. Human results based on a doubly annotated subset of the test set. Baseline results based on assigning salience scores that are iid randomly sampled or monotonically increasing/decreasing for each sentence. LLM prompting results using two strategies: same instructions as for human judgments; instruction to assign each sentence a global salience score. In bold are the winning results in each group according to permutation-based significance tests (see main text).}
\label{tab:roc-baseline}
\begin{small}
    \centering
    \begin{tabular}{lrr}
    \toprule
         & \multicolumn{1}{c}{$\bm{\rho}$} & \multicolumn{1}{c}{\textbf{AUC}} \\ \midrule
         \textsc{Human} & 0.50\pmzerofourspace & 0.74\pmzerofourspace \\ \midrule
         \textsc{Random} & 0.00$_{\pm0.03}$ & 0.50$_{\pm0.02}$ \\
         \textsc{Increasing} & \textbf{0.29}\pmzerofourspace & \textbf{0.65}\pmzerofourspace \\
         \textsc{Decreasing} & -0.29\pmzerofourspace &  0.35\pmzerofourspace \\ \midrule
         \multicolumn{3}{c}{annotator-style prompting} \\ \midrule
         \textsc{GPT-4.1 mini} & \textbf{0.51$_{\pm0.01}$} & \textbf{0.71$_{\pm0.00}$}\\
         \textsc{GPT-4.1} & \textbf{0.53$_{\pm0.00}$} &\textbf{0.71$_{\pm0.00}$} \\ \midrule
         \multicolumn{3}{c}{salience-style prompting} \\ \midrule
         \textsc{GPT-4.1 mini} & 0.45$_{\pm0.01}$ & 0.75$_{\pm0.00}$\\
         \textsc{GPT-4.1} & \textbf{0.52$_{\pm0.01}$} & \textbf{0.79$_{\pm0.01}$}\\
    \bottomrule
    \end{tabular}
   
\end{small}
\end{table}

\subsection{Training}
\label{sec:training}

The embedding model $f_\theta$ is a Transformer initialized with the weights of BERT \citep{devlin-2019-bert}.
We train on all samples in our training corpus for five epochs, using a batch size of 128 and a learning rate of 3e-5.
We use the AdamW optimizer \citep{loschchilov-2019-adamw} and for the contrastive loss we use a temperature of 0.05.
The Transformer generates as output contextualized embeddings for each token in the input.
To obtain narrative embeddings, we apply a mean pooling operation to this output.

We compare three approaches. 
The first continues the pre-training objective of masked language modeling on our training set (including all twins and distractors). 
The second uses our proposed model, with dropout twins as positive pairs.
The third uses the same model but with narrative twins as positive pairs.
Crucially, all three approaches are trained based on exactly the same data.

\subsection{Baselines and LLM Prompting}
\label{sec:baselines}

We provide three naive baselines: (1)~assigning a random value as the saliency of each sentence, (2)~assigning the sentences monotonically increasing saliency, and (3)~assigning them monotonically decreasing saliency.
As an upper bound, we provide the performance of a state-of-the-art LLM, using two different approaches to prompt it.
The first approach directly follows how our evaluation data was annotated: we provided the full annotation guidelines (with its two examples) and asked for a single label of the most important sentence.
We repeat the prompt (with a temperature of 1.0) ten times for each story and derive the ranking in the same way as performed for the evaluation data.
The second approach is to directly prompt the model to assign a saliency score to each sentence.
In the prompt, we defined saliency as the likelihood that information in a sentence would be included by someone else telling the same story.

The performance of these approaches is provided in Table~\ref{tab:roc-baseline}.
Random performance is $\rho=0.00$ and AUC = 0.50. Unsurprisingly, the baseline of salience increasing monotonically results in better performance ($\rho=0.29$ and AUC = 0.69);
clearly there is a bias in our data of the most salient sentences occurring towards the end of stories.
In terms of prompted LLMs, both approaches perform well.
Given that the first approach closely mimics the human annotation procedure, we expected it to perform best.
However, we find that for the best performing LLM (\textsc{GPT-4.1}) the second approach is either statistically indistinguishable or better.
The mean perplexity of the LLM labels from the first prompting method is 1.33 and 1.18 (for \textsc{GPT-4.1 mini} and \textsc{GPT-4.1}), a large reduction from the mean perplexity of the human labels (2.03).
The decreased variation in the labels may make this prompting method less suitable here.

\subsection{Results}
\label{sec:results}

Results for the three main training methods and four operations are in Table~\ref{tab:roc-main}.
For every operation, using narrative twins results in performance that is statistically better (shifting, disruption, summarization) or indistinguishable from the next best (deletion).
The overall winning operation is the summarization test ($\rho=0.44$, AUC = 0.76).
This result is significantly better or statistically indistinguishable from using \textsc{GPT-4.1 mini} in the two prompting setups ($\rho$: $p=0.062$, $p=0.670$; AUC: $p=0.006$, $p=0.957$).
In all cases it is significantly worse than prompting \textsc{GPT-4.1}.

The next-best operation is deletion, followed by shifting, and last is disruption. 
Appendix Figure~\ref{fig:rocstory-val} shows the validation performance at checkpoints throughout training. 
For the methods using twins, performance increases monotonically for all operations.
Masked LM either leads to small improvements, or performance is constant. In terms of qualitative results, Appendix~\ref{sec:qualitative_examples} shows the salience scores predicted by each operation on the two stories in our guidelines (the first one is Example~\ref{ex:roc-example}).

A requirement for good performance is a large enough (by $|\theta|$) embedding model: Appendix Figure~\ref{fig:rocstory-scaling} shows validation performance for the operations increasing with increased embedding model size \cite[we used the BERT tiny, mini, small, medium, base and large models,][]{turc-2019-well, devlin-2019-bert}.
Appendix Table~\ref{tab:roc-ablation} shows the results of four ablations: (1)~using the CLS token instead of mean pooling, (2)~training without negative distractors, (3)~substituting entities in the positive narrative twins according to \citet{hatzel-2024-story}, (4)~directly applying the operations to the pre-trained model without further training.

\section{Experiment 2: Wikipedia Narratives}
\label{sec:wikipedia}

We now apply our model to longer-form narratives, using our proposed model extension to windows (see Section~\ref{sec:model}). 
We looked for a narrative domain with an established benchmark related to predicting saliency.
A frequently investigated domain is Wikipedia summaries of movie plots.
One reason that they are an attractive domain for research is that they are written to conform to a standard template; hence they are uniform in terms of structure, style and length.
We use \citeposs{papalampidi-2019-movie} dataset of 99 Wikipedia narratives: in each summary, five sentences are annotated as turning points. We consider these to be narratively most salient.

\begin{table*}
\centering
\caption{Results on salience detection in Wikipedia narratives. All results are AUC scores averaged across five windows per story. In bold are the best performing approaches in each group, according to permutation-based significance tests. We report mean $\pm$ one standard deviation (computed across three different random seeds). In bold are the winning results in each group according to permutation-based significance tests (see main text).
}
\label{tab:tripod-main-paper}
\begin{small}
\begin{subtable}[b]{0.3\textwidth}
\centering
\caption{Baselines and LLM prompting}
\label{tab:tripod-baselines}
\begin{tabular}{lr}
\toprule
&  \multicolumn{1}{c}{\textbf{avg. AUC}} \\ \midrule
\textsc{Random}      & 0.50$_{\pm0.02}$ \\
\textsc{Increasing}  & \textbf{0.56}\pmzerofourspace \\
\textsc{Decreasing}  & 0.44\pmzerofourspace \\ \midrule
\textsc{GPT-4.1 mini}& 0.70$_{\pm0.01}$ \\
\textsc{GPT-4.1}     & \textbf{0.76$_{\pm0.00}$} \\
\textsc{GPT-5 mini}  & 0.72$_{\pm0.01}$ \\
\textsc{GPT-5}       & \textbf{0.76$_{\pm0.00}$} \\
\bottomrule
\end{tabular}
\end{subtable}
\hspace{0.05\textwidth}
\begin{subtable}[b]{0.6\textwidth}
\centering
\caption{Main results}
\label{tab:tripod-main-operations}
\begin{tabular}{lrrrr}
\toprule
& \multicolumn{1}{c}{\textbf{Deletion}} & \multicolumn{1}{c}{\textbf{Shifting}} & \multicolumn{1}{c}{\textbf{Disruption}} & \multicolumn{1}{c}{\textbf{Summariz.}} \\
\cmidrule(lr){2-2} \cmidrule(lr){3-3} \cmidrule(lr){4-4} \cmidrule(lr){5-5} 
&  \multicolumn{1}{c}{\textbf{avg. AUC}} &  \multicolumn{1}{c}{\textbf{avg. AUC}} &  \multicolumn{1}{c}{\textbf{avg. AUC}} &  \multicolumn{1}{c}{\textbf{avg. AUC}} \\ \midrule
\multicolumn{5}{c}{\textbf{LLM-negatives + in-story negatives}} \\ \midrule
\textsc{Masked LM}  & 0.56$_{\pm0.00}$ & 0.54$_{\pm0.00}$ & 0.49$_{\pm0.00}$ & 0.60$_{\pm0.00}$ \\
\textsc{Dropout Twins}    & 0.58$_{\pm0.00}$ & 0.58$_{\pm0.00}$ & 0.53$_{\pm0.00}$ & 0.63$_{\pm0.01}$ \\
\textsc{Narrative Twins}  & 0.60$_{\pm0.01}$ & 0.60$_{\pm0.00}$ & 0.54$_{\pm0.00}$ & \textbf{0.65$_{\pm0.00}$} \\ \midrule
\multicolumn{5}{c}{\textbf{Only in-story negatives}} \\ \midrule
\textsc{Masked LM} & 0.56$_{\pm0.00}$  & 0.54$_{\pm0.00}$ & 0.49$_{\pm0.00}$ & 0.59$_{\pm0.00}$ \\
\textsc{Dropout Twins} & 0.56$_{\pm0.00}$ & 0.57$_{\pm0.00}$ & 0.53$_{\pm0.00}$ & \textbf{0.65$_{\pm0.00}$} \\
\textsc{Narrative Twins} & 0.60$_{\pm0.01}$ & 0.60$_{\pm0.01}$ & 0.54$_{\pm0.00}$ & \textbf{0.66$_{\pm0.00}$} \\
\bottomrule
\end{tabular}
\end{subtable}

\end{small}
\end{table*}

\subsection{Narrative Twins and Negatives}
\label{sec:wiki:twins}

Rather than generating retellings using an LLM as in Experiment~1, we obtain narrative twins from \citeposs{hatzel-2024-tell} dataset (see Section~\ref{sec:embeddings}).
We use twins where the English Wikipedia narrative is of length greater than or equal to 20 sentences and the non-English Wikipedia narrative is of length within one standard deviation (13.7 sentences) of the English one.
If more than one twin is available for the same narrative, we randomly select one.

We align the English Wikipedia sentence sequence $A = (a_1,\dots,a_n)$ to the twin sequence $B = (b_1,\dots,b_m)$ via dynamic time warping over sentence similarities. 
We use the sentence encoder from \citet{gao-2021-sim_cse}.
Let $s_{ij}$ be the cosine similarity between $a_i$ and $b_j$.
We obtain the optimal monotonic path $P^\star \;=\; \arg\max_{P}\;\sum_{(i,j)\in P} s_{ij}$.
We compute this exactly in $O(nm)$ time with dynamic programming, as per the established protocol for this kind of alignment.

We split the Wikipedia sentence sequence into five evenly spaced windows (each window is thus $\geq4$ sentences).
The alignment $P^\star$ provides the aligned windows in the twin.
We only keep twins for which every aligned window is $\geq3$ sentences.
We include each source narrative in an LLM (\textsc{GPT-5 mini}) prompt and obtain a negative distractor.
In our prompt, we are prescriptive about each of the turning points and tell the model to change the turning points and hence underlying story.
Meanwhile, the LLM is instructed to keep all of the same characters, settings, and to integrate the existing minor events as a logical part of the new story.
For a test of one of our hypotheses, we will compare performance using these LLM hard negatives with and without also using in-story negatives (see Section~\ref{sec:model}).
The LLM prompt we use is in Appendix~\ref{sec:wik:prompts}.
In total, we have 2.2k narrative twins and negatives.

\subsection{Training}
\label{sec:wki:training}

The Wikipedia narratives are much longer than the earlier ROCStories (mean length $33.9\pm9.1$ sentences).
As a base Transformer pre-trained to handle sufficiently long context, we use ModernBERT \citep{warner-2025-smarter}.
The training procedure of this model involves sequences of length 1,024 and subsequently 8,192 tokens.
We use the 1B parameter pre-trained ModernBERT model from \citet{weller-2025-seqvsseq}.
The model takes each complete Wikipedia narrative as input.
A representation of each window in the narrative is obtained by mean pooling the tokens associated with the window. 
The training objective is for a window embedding to be similar to the window embedding in the twin, and different from every other window representation in the twin (if in-story negatives are used) and every other in-batch window representation.
We use the same training parameters as for our earlier experiment on ROCStories.
The only difference is that we reduce the batch size to~16, or~4 for ablation runs.\footnote{All main experiments were performed on a single H200 GPU with 141GB of RAM, and ablations were performed on a single L40S GPU. For the long-form narrative experiments, we used flash attention v2.}

\subsection{Metrics}
\label{sec:wiki:metrics}

For evaluation, we only have access to the single-sentence labels of the turning points, so ranking-based $\rho$ cannot be used.
On average, the five labels are spread evenly throughout the narratives.
Following \citeposs{papalampidi-2019-movie} evaluation protocol, we thus treat each window as having exactly one most salient sentence (the corresponding turning point). 
Rather than giving a score of zero to all models, we simply ignore windows where each label is outside the window (29\% of windows).
For the remaining windows, we compare system saliency scores with the turning point label using~AUC.
Since there is only one label, AUC corresponds to the proportion of sentences in the window predicted to be less salient than the labeled sentence divided by the total number of sentences in the window.

\subsection{Baselines and LLM Prompting}
\label{sec:wiki:baselines}

We provide the same three baselines as our earlier experiments on ROCStories: random, increasing and decreasing.
Given the effectiveness of the earlier LLM salience-style prompting, we repeat it here to obtain an upper limit.
We use an identical prompt, with the only difference that the number of sentences in the input is longer and can vary.

Results are shown in Table~\ref{tab:tripod-baselines}.
Increasing saliency is the best baseline (AUC = 0.56).
The prompted LLMs perform better (best AUC = 0.76).
There is no improvement between comparable versions of \textsc{GPT-4.1} and \textsc{GPT-5}; in both cases the largest model performs better than \textsc{mini} (AUC +0.06; +0.04).

\subsection{Results}
\label{sec:wiki:results}

Our main results are given in Table~\ref{tab:tripod-main-operations}.
The narrative operation that is best predicting saliency is again summarization (for every row, differences between summarization and other operations are significant, $p<0.05$).
We observe no statistical difference between using both LLM-generated negatives together with in-story negatives and using only in-story negatives (except dropout twins with summarization for which adding the LLM negatives leads to a significant reduction in performance, $p=0.040$) .
In both cases, for the summarization operation, dropout twins and narrative twins are also statistically indistinguishable.
This is evidence in support of our hypothesis that dropout twins can be used to generate narrative twins at the representation level.
Furthermore, the success of the in-story negatives supports our hypothesis that this approach can serve as a method of adding hard negatives into the contrastive learning on long-form narratives.
Indeed with dropout twins and only in-story negatives we have achieved our joint-best results in a fully self-supervised setup.

The joint second-best operations are deletion and shifting, which are indistinguishable for both twin approaches.
Meanwhile the disruption operation is again significantly worse.
This is the same ordering of the performance of the operations as Experiment 1 (modulo deletion and shifting being distinguishable on ROCStories, but not here).
For all operations, using narrative twins is significantly better than masked LM training (all $p<0.01$).
This result shows that the contrastive training is providing a learning signal that is useful to the task.

We performed two ablations: One in which we remove the negative distractors from the contrastive loss, and another where we additionally use \citeauthor{hatzel-2024-story}'s entity-substitution strategy (see Appendix~\ref{sec:comp:results:wiki} for details).
These ablations show performance degradations, which aligns with our expectation that negative distractors are vital for the contrastive learning on long-form narratives.
Appendix~\ref{sec:qualitative_examples} shows the salience scores predicted by each of the operations on narrative~\ref{ex:dragon-tattoo}, from \textit{The Girl With A Dragon Tattoo}.

\section{Conclusions and Future Work}
\label{sec:conclusions}

In this work, we showed how to train story embeddings using narrative twins. We demonstrated that these embeddings perform well in the task of determining what is salient in a story. This was complemented by an investigation of which operationalizations of salience proposed in narratology perform best. We found that the summarization test works best both on five-sentence narratives (ROCStories) and on longer narratives (Wikipedia plot summaries). This test selects the sentence that best summarizes the story, and can be computed by comparing the embedding of each sentence with the embedding of the whole narrative.

Our experiments also showed that if narrative twins are not available, dropout twins are a good substitute. Dropout twins are computed automatically from the original narrative, because they constitute two embeddings of the same story that differ because different random dropout values have been used in the embedding model.

In both of our experiments, the annotations we used for narrative salience were at the sentence-level.
However, a sentence may contain more than one event, and those events will frequently have different salience.
Narrative salience is also hierarchical \citep{barthes-1966-introduction, cohn-2013-visual}: climactic chapters are typically more salient than other chapters, and narratives have high-level themes that are salient.
Further experimentation would require new annotation at these levels.
This connects to future work on training a model contrastively at different levels (e.g.,~sentences as well as windows).

The longest-form narratives we explored in this work were Wikipedia narratives, which are on average 34 sentences long.
Meanwhile many real-world narratives are much longer.
We expect that with increasing length it will not be possible to rely on LLMs for labels of global scores of narrative salience.
Instead, the application of the model and operations presented in this work is more feasible. 

\section*{Limitations}
\label{sec:limitations}

We have shown that the performance of the narrative operations depend on how the embeddings are generated.
We aimed for a controlled comparison between the operations, but our results may only hold for the methodologies that we have tested here for generating the embeddings.

Furthermore, a component required for the success of our model is the use of negative distractor narratives.
For our experiments with short-form narratives, the only approach we proposed to obtain such distractors was by prompting LLMs.
Our prompt explicitly stated that the LLM should prioritize changing important events.
In this way, here we have injected a form of weak LLM supervision into our model.

\section*{Ethical Considerations}
\label{sec:ethics}

Our experiment with human participants was approved by the departmental ethics committee.
This included consideration of gaining informed consent from participants and the amount of payment they received for their time.
We paid all participants equally, regardless of whether they passed the comprehension check or not.

We train on Wikipedia narratives from \citet{hatzel-2024-tell}, which are available under a permissive license and designated for model training.
We have released all code and data required to reproduce our experimental results (see Footnote~\ref{fn:data}). Note that the summaries have been released alongside the salience labels under the same permissive license as the ROCStories. 

\section*{Acknowledgements}

IS is funded by an EPSRC PhD studentship (project reference 2923920).
Annotation on the ROCStories was funded by the Institute for Language, Cognition and Computation, School of Informatics, University of Edinburgh.
We thank Sydelle de Souza for help in designing the ROCStory human experiment, and Inderjeet Mani for comments.

\bibliography{myrefs}

\appendix

\section{Complementary Results}
\label{sec:comp:results}

\subsection{ROCStories}
\label{sec:comp:results:roc}

\begin{figure*}[h]
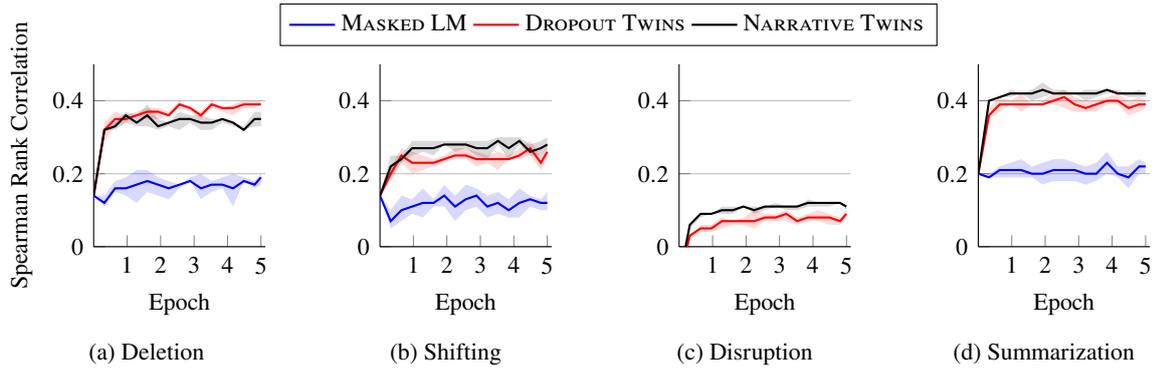

    
    \begin{subfigure}[b]{0.24\textwidth}
    \centering
        \begin{tikzpicture}
        \begin{axis}[
            xmin=0, xmax=800,
            ymin=0.0, ymax=0.5,
            ymajorgrids=true,
            xlabel=Epoch,
            ylabel=Spearman Rank Correlation,
            height=4cm,
            width=\linewidth,
            ylabel style={font=\small},
            xlabel style={font=\small},
            xticklabel style={font=\small},
            yticklabel style={font=\small},
            xtick={156, 312, 468, 624, 780},
            xticklabels={1, 2, 3, 4, 5},
            axis x line*=bottom,
            axis y line*=left,
            legend style={
                at={(3,1.1)},
                anchor=south,
                legend columns=3,
                font=\small,
                align=left,
                cells={anchor=west},
            },
        ]
        \input{data/rocstories/plots/deletion}

        \end{axis}
        \end{tikzpicture}
    \caption{Deletion}
    \label{fig:del-graph}
    \end{subfigure}
    \begin{subfigure}[b]{0.24\textwidth}
    \centering
        \begin{tikzpicture}
        \begin{axis}[
            xmin=0, xmax=800,
            ymin=0.0, ymax=0.5,
            ymajorgrids=true,
            xlabel=Epoch,
            height=4cm,
            width=\linewidth,
            ylabel style={font=\small},
            xlabel style={font=\small},
            xticklabel style={font=\small},
            yticklabel style={font=\small},
            xtick={156, 312, 468, 624, 780},
            xticklabels={1, 2, 3, 4, 5},
            axis x line*=bottom,
            axis y line*=left,
            legend style={
                at={(0.3,1.1)},
                anchor=south,
                legend columns=2,
                font=\small,
                align=left,
                cells={anchor=west},
            },
        ]

        \input{data/rocstories/plots/shifting}
        \end{axis}
        \end{tikzpicture}
    \caption{Shifting}
    \label{fig:shi-graph}
    \end{subfigure}
    \begin{subfigure}[b]{0.24\textwidth}
    \centering
        \begin{tikzpicture}
        \begin{axis}[
            xmin=0, xmax=800,
            ymin=0.0, ymax=0.5,
            ymajorgrids=true,
            xlabel=Epoch,
            height=4cm,
            width=\linewidth,
            ylabel style={font=\small},
            xlabel style={font=\small},
            xticklabel style={font=\small},
            yticklabel style={font=\small},
            xtick={156, 312, 468, 624, 780}, xticklabels={1, 2, 3, 4, 5},
            axis x line*=bottom,
            axis y line*=left,
            legend style={
                at={(0.3,1.1)},
                anchor=south,
                legend columns=3,
                font=\small,
                align=left,
                cells={anchor=west},
            },
        ]

        \input{data/rocstories/plots/disruption}
        \end{axis}
        \end{tikzpicture}
    \caption{Disruption}
    \label{fig:dis-graph}
    \end{subfigure}
    \begin{subfigure}[b]{0.24\textwidth}
    \centering
        \begin{tikzpicture}
        \begin{axis}[
            xmin=0, xmax=800,
            ymin=0.0, ymax=0.5,
            ymajorgrids=true,
            xlabel=Epoch,
            height=4cm,
            width=\linewidth,
            ylabel style={font=\small},
            xlabel style={font=\small},
            xticklabel style={font=\small},
            yticklabel style={font=\small},
            xtick={156, 312, 468, 624, 780}, 
            xticklabels={1, 2, 3, 4, 5},
            axis x line*=bottom,
            axis y line*=left,
            legend style={
                at={(0.3,1.1)},
                anchor=south,
                legend columns=2,
                font=\small,
                align=left,
                cells={anchor=west},
            },
        ]
        \input{data/rocstories/plots/summarisation}
        \end{axis}
        \end{tikzpicture}
    \caption{Summarization}
    \label{fig:sum-graph}
    \end{subfigure}
\caption{Validation performance on ROCStories throughout training. Curves show Spearman rank correlation (which has range [-1, 1]; random is 0.0; higher is better) between model-predicted salience and human rankings. All models are trained with the same data. In blue is a model trained with masked language modeling. In red is a model trained with contrastive learning and positives derived from applying different random dropout on the same input. Black line represents model trained with textual narrative twins.}
\label{fig:rocstory-val}
\end{figure*}

Figure~\ref{fig:rocstory-val} shows performance on our ROCStory validation set on a model checkpoint each 50 steps of training.
(Note that with masked language modelling, the disruption operation performs with $\rho<0$ (i.e., the rankings are negatively correlated) and hence the line does not appear in the plot.)
For all operations and both types of narrative twins, we observe monotonically increasing performance. 
Masked language modeling offers a form of domain-adaptation, but the plots here confirm that this does not lead to improved performance on our task.

Results of our ROCStory ablation runs are in Table~\ref{tab:roc-ablation}.
As our first ablation, we compared using the model's first token (the CLS token) as the narrative embedding, rather than mean-pooling all token embeddings.
The motivation for this ablation is that language models use layer-wise normalization \citep{ba-2016-layernorm}.
One result of this is that the output token representations is an affine transform of a normalized vector, which could make the task harder for a model trained to learn to represent salient events more than less-salient events. 
The ablation shows no difference in performance with using first token as compared to mean pooling.
In a second ablation, we remove the negative distractors from the batches.
Results for this show drastically reduced performance (for narrative twins and summarization, $\rho$=0.76 $\rightarrow$ 0.60), which supports our intuition that hard negative distractors are vital for the learning.
In a final ablation, we tried training the model without the negative distractors, but substituting entities in the positive narrative twins according to \citeposs{hatzel-2024-story} strategy.
This involves applying coreference resolution and named entity recognition models.\footnote{We adapt code released by \citet{hatzel-2024-story}.} 
As an example, \ref{ex:roc-example-anonymized} shows the result for the earlier~\ref{ex:roc-example} (without this intervention the twin is as per \ref{ex:roc-example-twin} (page \pageref{ex:roc-example-twin})).
The change is that Bob has been renamed Robert, and Ricky has been renamed Dennis.
Note this strategy also substitutes location and organization names, but none are present in this example.
Results show that this strategy performs better than the previous ablation, but performance is still worse than using the hard negative distractors.
We expect that the presence of other shallow overlap in a narrative twin (e.g., topical overlap) remains a problematic shortcut.

\ex.\label{ex:roc-example-anonymized}
\a. Dennis and Robert decided to play tennis one sunny afternoon.
\b. Robert served the ball, but it accidentally hit Dennis on the head, and he fainted.
\c. Robert quickly checked on Dennis and poured water into his mouth.
\d. Dennis woke up and felt better after resting.
\e. They continued to play, being more cautious this time.

\begin{table*}[h]
\centering
\caption{ROCStory ablation study results.}
\label{tab:roc-ablation}
\begin{small}
\begin{tabular}{lrrrrrrrr}
\toprule
& \multicolumn{2}{c}{\textbf{Deletion}} & \multicolumn{2}{c}{\textbf{Shifting}} & \multicolumn{2}{c}{\textbf{Disruption}} & \multicolumn{2}{c}{\textbf{Summarization}} \\
\cmidrule(lr){2-3} \cmidrule(lr){4-5} \cmidrule(lr){6-7} \cmidrule(lr){8-9} 
& \multicolumn{1}{c}{\textbf{$\bm{\rho}$}} & \multicolumn{1}{c}{\textbf{AUC}} & \multicolumn{1}{c}{\textbf{$\bm{\rho}$}} & \multicolumn{1}{c}{\textbf{AUC}} & \multicolumn{1}{c}{\textbf{$\bm{\rho}$}} & \multicolumn{1}{c}{\textbf{AUC}} & \multicolumn{1}{c}{\textbf{$\bm{\rho}$}} & \multicolumn{1}{c}{\textbf{AUC}} \\ \midrule
\multicolumn{9}{c}{\textbf{Base}} \\ \midrule
\textsc{Masked LM} & 0.20$_{\pm0.01}$ & 0.61$_{\pm0.01}$ & 0.17$_{\pm0.04}$ & 0.59$_{\pm0.02}$ & -0.14$_{\pm0.00}$ & 0.42$_{\pm0.00}$ & 0.21$_{\pm0.01}$ & 0.62$_{\pm0.01}$ \\
\textsc{Dropout Twins} & 0.37$_{\pm0.01}$ & 0.70$_{\pm0.01}$ & 0.25$_{\pm0.01}$ & 0.63$_{\pm0.00}$ & 0.06$_{\pm0.01}$ & 0.53$_{\pm0.01}$ & 0.35$_{\pm0.01}$ & 0.70$_{\pm0.00}$ \\
\textsc{Narrative Twins} & 0.38$_{\pm0.00}$ & 0.70$_{\pm0.00}$ & 0.31$_{\pm0.01}$ & 0.66$_{\pm0.00}$ & 0.13$_{\pm0.00}$ & 0.57$_{\pm0.00}$ & 0.44$_{\pm0.01}$ & 0.76$_{\pm0.00}$ \\ \midrule
\multicolumn{9}{c}{\textbf{CLS token}} \\ \midrule
\textsc{Dropout Twins} & 0.38$_{\pm0.01}$ & 0.70$_{\pm0.00}$ & 0.25$_{\pm0.00}$ & 0.62$_{\pm0.00}$ & 0.04$_{\pm0.00}$ & 0.52$_{\pm0.00}$ & 0.40$_{\pm0.00}$ & 0.73$_{\pm0.00}$ \\
\textsc{Narrative Twins} & 0.38$_{\pm0.01}$ & 0.71$_{\pm0.00}$ & 0.33$_{\pm0.01}$ & 0.67$_{\pm0.00}$ & 0.13$_{\pm0.00}$ & 0.57$_{\pm0.00}$ & 0.44$_{\pm0.01}$ & 0.75$_{\pm0.01}$ \\ \midrule
\multicolumn{9}{c}{\textbf{without hard negative distractors}} \\ \midrule
\textsc{Dropout Twins} & 0.05$_{\pm0.00}$ & 0.53$_{\pm0.00}$ & 0.09$_{\pm0.02}$ & 0.56$_{\pm0.01}$ & -0.09$_{\pm0.01}$ & 0.45$_{\pm0.00}$ & 0.14$_{\pm0.02}$ & 0.58$_{\pm0.01}$ \\
\textsc{Narrative Twins} & 0.03$_{\pm0.02}$ & 0.52$_{\pm0.01}$ & 0.06$_{\pm0.00}$ & 0.54$_{\pm0.00}$ & -0.07$_{\pm0.01}$ & 0.46$_{\pm0.01}$ & 0.16$_{\pm0.01}$ & 0.60$_{\pm0.00}$ \\ \midrule
\multicolumn{9}{c}{\textbf{without hard negative distractors; with entity-substitution}} \\ \midrule
\textsc{Dropout Twins} & 0.05$_{\pm0.00}$ & 0.53$_{\pm0.00}$ & 0.07$_{\pm0.00}$ & 0.55$_{\pm0.00}$ & -0.09$_{\pm0.01}$ & 0.45$_{\pm0.00}$ & 0.15$_{\pm0.02}$ & 0.59$_{\pm0.01}$ \\
\textsc{Narrative Twins} & 0.17$_{\pm0.02}$ & 0.59$_{\pm0.01}$ & 0.13$_{\pm0.01}$ & 0.57$_{\pm0.01}$ & -0.03$_{\pm0.01}$ & 0.48$_{\pm0.01}$ & 0.27$_{\pm0.01}$ & 0.66$_{\pm0.01}$ \\ \midrule
\multicolumn{9}{c}{\textbf{no Training}} \\ \midrule
\textsc{CLS token} & 0.27\pmzerofourspace & 0.64\pmzerofourspace & 0.17\pmzerofourspace & 0.57\pmzerofourspace & -0.02\pmzerofourspace & 0.49\pmzerofourspace & 0.21\pmzerofourspace & 0.62\pmzerofourspace \\
\textsc{Mean pooling} & 0.15\pmzerofourspace & 0.58\pmzerofourspace & 0.09\pmzerofourspace & 0.54\pmzerofourspace & -0.12\pmzerofourspace & 0.43\pmzerofourspace & 0.19\pmzerofourspace & 0.61\pmzerofourspace \\ 
\bottomrule
\end{tabular}
\end{small}
\end{table*}

\begin{figure*}[h]
    \centering
    \begin{tikzpicture}
\begin{axis}[
    xtick style={draw=none},
    xticklabel={\pgfmathprintnumber{\tick}M},
    ytick pos=left,
    ymin=0,
    ymax=0.5,
    ymode=linear,
    scaled y ticks=false,
    xmin=4.43,
    xmax=335.14,
    ymajorgrids=true,
    grid style=dashed,
	xlabel=\# model parameters,
	ylabel=$\rho$,
    ylabel style={font=\small},
    xlabel style={font=\small},
    xticklabel style={font=\small},
    yticklabel style={font=\small},
	width=0.4\linewidth,
    height=3.5cm,
    legend style={at={(1.2,0.5)},anchor=west, font=\small, legend cell align={left}}
]

% Deletion
\addplot[color=red, line width=1pt] table  {
Size	Deletion
4.43	0.077
11.17	0.249
28.76	0.240
41.37	0.350
109.48	0.350
335.14	0.349
};
% Shifting
\addplot[color=blue, line width=1pt] table  {
Size	Shifting
4.43	0.018
11.17	0.037
28.76	0.122
41.37	0.157
109.48	0.220
335.14	0.270
};
% Disruption
\addplot[color=black, line width=1pt] table  {
Size	Disruption
4.43	-0.07
11.17	0.010
28.76	0.0413
41.37	0.049
109.48	0.068
335.14	0.103
};
% Summarisation
\addplot[color=green, line width=1pt] table  {
Size	Summarisation
4.43	0.23655
11.17	0.317
28.76	0.385
41.37	0.407
109.48	0.446
335.14	0.411
};

\legend{Deletion, Shifting, Disruption, Summarisation}

\end{axis}

\end{tikzpicture}
    \caption{ROCStory model performance with increasing model size.}
    \label{fig:rocstory-scaling}
\end{figure*}
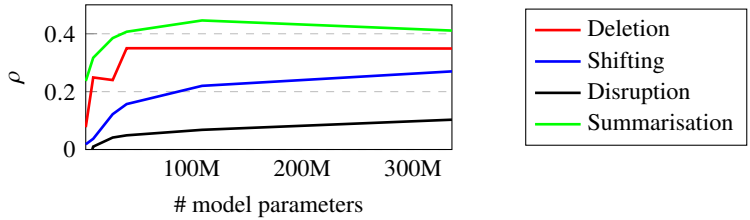

Figure~\ref{fig:rocstory-scaling} shows performance for each of the narrative operations with increasing model size.
We performed runs with the base model initialized with BERT tiny, mini, small, medium, base and large \citep{turc-2019-well, devlin-2019-bert}.
We can see that even for these short narratives, a large-enough model is required for good performance.

\subsection{Wikipedia Narratives}
\label{sec:comp:results:wiki}

Table~\ref{tab:tripod-results-full} provides a breakdown of Wikipedia narrative results by each of the five turning point (TP) types as defined in \citet{papalampidi-2019-movie}: \textit{Opportunity} (TP1), \textit{Change of Plans} (TP2), \textit{Point of No Return} (TP3), \textit{Major Setback} (TP4), and \textit{Climax} (TP5).
The results show that accuracy varies both by turning point type and by the narrative operation applied.
This suggests that certain operations are more effective for detecting particular turning points.
For instance, the disruption test is the best operation for detecting the \textit{Opportunity} turning point.
This is intuitive, because the presentation of an opportunity dramatically changes the story's interpretation.
Meanwhile, across most of the models the salience of the \textit{Climax} turning point is predicted with highest accuracy.
This is especially true for the prompted LLMs.
Of the narrative operations, summarization is the most consistent across all turning points.

Table~\ref{tab:wikipedia-ablation} gives results of 2 ablation runs on Wikipedia narrative, as well as the performance of the base model without further finetuning.
Here, we use the results with in-story negatives as base performance.
In the first ablation, we remove the in-story negatives from the contrastive comparison.
This leads to performance drops.
In our second ablation, we use \citeposs{hatzel-2024-story} strategy of substituting entities in the narrative twins.
For this we directly use \citeposs{hatzel-2024-tell} data with substituted entities, recomputing the alignment required for our model.
Results show either zero or modest improvement compared to without the substitution.
(Note that these results are still based on applying the contrastive learning at the window-level, which is different to the narrative-level contrastive loss used by \citet{hatzel-2024-story}.)
It is evident that \citeposs{hatzel-2024-story} strategy of dealing with the lexical overlap does contribute to solving the challenge, but our methods show better results.

\begin{table*}[t]
\centering
\caption{Full results on salience detection in Wikipedia narratives. Results here are broken down by each turning point label. TP1-5 correspond to labels of \textit{Opportunity}, \textit{Change of Plans}, \textit{Point of No Return}, \textit{Major Setback} and \textit{Climax} (see \citet{papalampidi-2019-movie} for more details).}
\label{tab:tripod-results-full}
\begin{small}
\begin{tabular}{lrrrrrr}
\toprule
 & \multicolumn{1}{c}{TP1} &
   \multicolumn{1}{c}{TP2} &
   \multicolumn{1}{c}{TP3} &
   \multicolumn{1}{c}{TP4} &
   \multicolumn{1}{c}{TP5} &
   \multicolumn{1}{c}{\textbf{avg.}} \\ \midrule
\multicolumn{7}{c}{\textbf{Baselines}} \\ \midrule
\textsc{Random} & 0.50$_{\pm0.04}$ & 0.51$_{\pm0.05}$ & 0.50$_{\pm0.04}$ & 0.50$_{\pm0.04}$ & 0.49$_{\pm0.04}$ & 0.50$_{\pm0.02}$ \\ 
\textsc{GPT 5 Mini} & 0.73$_{\pm0.00}$  & 0.69$_{\pm0.01}$  & 0.71$_{\pm0.02}$  & 0.65$_{\pm0.02}$  & 0.81$_{\pm0.01}$ & 0.72$_{\pm0.01}$  \\
\textsc{GPT 5} & 0.76$_{\pm0.01}$  & 0.77$_{\pm0.00}$  & 0.74$_{\pm0.01}$  & 0.69$_{\pm0.01}$  & 0.84$_{\pm0.01}$ & {0.76$_{\pm0.00}$}   \\ \midrule
\multicolumn{7}{c}{\textbf{Deletion}} \\ \midrule
\textsc{Masked LM} & 0.60$_{\pm0.00}$ & 0.58$_{\pm0.00}$ & 0.53$_{\pm0.00}$ & 0.52$_{\pm0.00}$ & 0.57$_{\pm0.00}$ & 0.56$_{\pm0.00}$ \\
\textsc{Dropout Twins} & 0.63$_{\pm0.01}$ & 0.56$_{\pm0.01}$ & 0.54$_{\pm0.00}$ & 0.55$_{\pm0.01}$ & 0.61$_{\pm0.01}$ & 0.58$_{\pm0.00}$ \\
\textsc{Narrative Twins} & 0.64$_{\pm0.01}$ & 0.58$_{\pm0.01}$ & 0.60$_{\pm0.01}$ & 0.54$_{\pm0.00}$ & 0.66$_{\pm0.00}$ & {0.60$_{\pm0.01}$} \\
\midrule
\multicolumn{7}{c}{\textbf{Shifting}} \\ \midrule
\textsc{Masked LM} & 0.57$_{\pm0.00}$ & 0.61$_{\pm0.00}$ & 0.53$_{\pm0.00}$ & 0.52$_{\pm0.00}$ & 0.50$_{\pm0.00}$ & 0.54$_{\pm0.00}$ \\
\textsc{Dropout Twins} & 0.58$_{\pm0.01}$ & 0.58$_{\pm0.01}$ & 0.59$_{\pm0.01}$ & 0.56$_{\pm0.01}$ & 0.59$_{\pm0.00}$ & 0.58$_{\pm0.00}$ \\
\textsc{Narrative Twins} & 0.60$_{\pm0.01}$ & 0.59$_{\pm0.01}$ & 0.59$_{\pm0.01}$ & 0.59$_{\pm0.00}$ & 0.62$_{\pm0.00}$ & {0.60$_{\pm0.00}$} \\
\midrule
\multicolumn{7}{c}{\textbf{Disruption}} \\ \midrule
\textsc{Masked LM} & 0.59$_{\pm0.00}$ & 0.57$_{\pm0.00}$ & 0.48$_{\pm0.00}$ & 0.37$_{\pm0.00}$ & 0.43$_{\pm0.00}$ & 0.49$_{\pm0.00}$ \\
\textsc{Dropout Twins} & 0.67$_{\pm0.01}$ & 0.59$_{\pm0.00}$ & 0.49$_{\pm0.00}$ & 0.44$_{\pm0.00}$ & 0.48$_{\pm0.00}$ & 0.53$_{\pm0.00}$ \\
\textsc{Narrative Twins} & 0.66$_{\pm0.01}$ & 0.54$_{\pm0.01}$ & 0.52$_{\pm0.00}$ & 0.43$_{\pm0.01}$ & 0.53$_{\pm0.00}$ & {0.54$_{\pm0.00}$} \\
\midrule
\multicolumn{7}{c}{\textbf{Summarization}} \\ \midrule
\textsc{Masked LM} & 0.61$_{\pm0.00}$ & 0.65$_{\pm0.00}$ & 0.52$_{\pm0.00}$ & 0.57$_{\pm0.00}$ & 0.63$_{\pm0.00}$ & 0.60$_{\pm0.00}$ \\
\textsc{Dropout Twins} & 0.63$_{\pm0.01}$ & 0.63$_{\pm0.00}$ & 0.64$_{\pm0.02}$ & 0.60$_{\pm0.02}$ & 0.67$_{\pm0.01}$ & 0.63$_{\pm0.01}$ \\
\textsc{Narrative Twins} & 0.63$_{\pm0.00}$ & 0.67$_{\pm0.00}$ & 0.63$_{\pm0.01}$ & 0.66$_{\pm0.01}$ & 0.65$_{\pm0.00}$ & {0.65$_{\pm0.00}$} \\
\bottomrule
\end{tabular}
\end{small}

\end{table*}

\begin{table*}[h]
\centering
\caption{Wikipedia narrative ablation study results.}
\label{tab:wikipedia-ablation}
\begin{small}
\begin{tabular}{lrrrr}
\toprule
& \multicolumn{1}{c} {\textbf{Deletion}} & \multicolumn{1}{c}{\textbf{Shifting}} & \multicolumn{1}{c}{\textbf{Disruption}} & \multicolumn{1}{c}{\textbf{Summariz.}} \\
\cmidrule(lr){2-2} \cmidrule(lr){3-3} \cmidrule(lr){4-4} \cmidrule(lr){5-5} 
 & \multicolumn{1}{c}{\textbf{avg. AUC}}  & \multicolumn{1}{c}{\textbf{avg. AUC}}  & \multicolumn{1}{c}{\textbf{avg. AUC}}  & \multicolumn{1}{c}{\textbf{avg. AUC}} \\ \midrule
\multicolumn{5}{c}{\textbf{Base}} \\ \midrule
\textsc{Dropout Twins} & 0.56$_{\pm0.00}$ & 0.57$_{\pm0.00}$ & 0.53$_{\pm0.00}$ & 0.65$_{\pm0.00}$ \\
\textsc{Narrative Twins} & 0.60$_{\pm0.01}$ & 0.60$_{\pm0.01}$ & 0.54$_{\pm0.00}$ & 0.66$_{\pm0.00}$ \\ 
\midrule
\multicolumn{5}{c}{\textbf{w/o hard negative distractors}} \\ \midrule
\textsc{Dropout Twins} & 0.56$_{\pm0.01}$ & 0.55$_{\pm0.01}$ & 0.51$_{\pm0.00}$ & 0.63$_{\pm0.01}$ \\
\textsc{Narrative Twins} & 0.56$_{\pm0.01}$ & 0.56$_{\pm0.01}$ & 0.51$_{\pm0.00}$ & 0.63$_{\pm0.01}$ \\ 
\midrule
\multicolumn{5}{c}{\textbf{w/o hard negative distractors; w/ entity substitution}} \\ \midrule
\textsc{Dropout Twins} & 0.55$_{\pm0.00}$ & 0.55$_{\pm0.00}$ & 0.50$_{\pm0.00}$ & 0.63$_{\pm0.00}$ \\
\textsc{Narrative Twins} & 0.56$_{\pm0.00}$ & 0.57$_{\pm0.01}$ & 0.52$_{\pm0.00}$ & 0.65$_{\pm0.00}$ \\ 
\midrule
\multicolumn{5}{c}{\textbf{no training}} \\ \midrule
\textsc{Window-Mean} & 0.56\pmzerofourspace & 0.54\pmzerofourspace & 0.49\pmzerofourspace & 0.59\pmzerofourspace\\
\bottomrule
\end{tabular}
\end{small}
\end{table*}

\begin{table*}
\centering
\caption{Predicted salience scores for each sentence in Story~\ref{ex:roc-example}.}
\label{tab:rocstories-qualitative-1}
\begin{small}
\begin{tabular}{lrrrr}
\toprule
\textbf{Sentence} & \textbf{Deletion} & \textbf{Shifting} & \textbf{Disruption} & \textbf{Summarization} \\ \midrule
Ricky was playing tennis with Bob. & 0.01773 & 0.03973 & 0.00000 & 0.35098 \\
Bob served the ball and hit Ricky right in the head! & 0.16117 & 0.08829 & 0.60952 & 0.59289 \\
Ricky fainted on the tennis court. & 0.20448 & 0.05167 & 0.25631 & 0.75795 \\
Bob gave Ricky some water to drink. & 0.05280 & 0.03074 & 0.07316 & 0.39018 \\
Ricky felt better after resting for awhile. & 0.00908 & 0.01072 & 0.00908 & 0.45127 \\
\bottomrule
\end{tabular}
\end{small}
\end{table*}

\begin{table*}[h]
\centering
\caption{Predicted salience scores for each sentence in Story~\ref{ex:my-youngest}.}
\label{tab:rocstories-qualitative-2}
\begin{small}
\begin{tabular}{lrrrr}
\toprule
\textbf{Sentence} & \textbf{Deletion} & \textbf{Shifting} & \textbf{Disruption} & \textbf{Summarization} \\ \midrule
My youngest nephew came to stay with us for a weekend. & 0.11267 & 0.08801 & 0.00000 & 0.46737 \\
He taught my youngest son lots of new words. & 0.03196 & 0.02425 & 0.38313 & 0.28453 \\
He showed him some next and interesting gestures. & 0.01182 & 0.01866 & 0.12434 & 0.19739 \\
Finally, he gave us all a tutorial on tantrum throwing. & 0.30280 & 0.05622 & 0.33144 & 0.74473 \\
I was elated to see his parents come and pick him up. & 0.07609 & 0.03857 & 0.07609 & 0.19373 \\
\bottomrule
\end{tabular}

\end{small}
\end{table*}

\section{Qualitative Examples}
\label{sec:qualitative_examples}

Table~\ref{tab:rocstories-qualitative-1} and \ref{tab:rocstories-qualitative-2} show the two ROCStories that we used in our guidelines.
Next to each sentence, we show the saliency score predicted for each sentence using each operation.

Table~\ref{tab:dragon-tattoo} shows each sentence in the aforementioned (ex.~\ref{ex:dragon-tattoo}, p.\pageref{ex:dragon-tattoo}) Wikipedia plot summary of \textit{The Girl with the Dragon Tattoo} (2011 film).
Next to each sentence is the saliency score predicted by each operation. 
Horizontal lines indicate the five evenly spaced windows.
Recall that the operations here are being applied at the window level.

{
\onecolumn
\begin{small}

\begin{longtable}{p{8cm}cccc}
\caption{Predicted salience scores for each sentence in the Wikipedia plot summary of \textit{The Girl with the Dragon Tattoo} (2011 film). In bold are the 5 sentences \citet{papalampidi-2019-movie} annoated as turning points; these are our labels of the most salient sentences.}\label{tab:dragon-tattoo} \\
\toprule
\textbf{Sentence} & \textbf{Deletion} & \textbf{Shifting} & \textbf{Disruption} & \textbf{Summariz.} \\ \midrule
In Stockholm, Sweden, journalist Mikael Blomkvist, co-owner of Millennium magazine, has just lost a libel case brought against him by businessman Hans-Erik Wennerström. & 0.058862 & 0.023842 & 0.000000 & 0.525191 \\
Meanwhile, Lisbeth Salander, a brilliant but troubled investigator and hacker, compiles an extensive background check on Blomkvist for business magnate Henrik Vanger, who has a special task for him. & 0.023475 & 0.015147 & 0.158102 & 0.601780 \\
\textbf{In exchange for the promise of damning information about Wennerström, Blomkvist agrees to investigate the disappearance and assumed murder of Henrik's grandniece, Harriet, 40 years ago.} & 0.035807 & 0.013807 & 0.046755 & 0.570716 \\
After moving to the Vanger family's compound, Blomkvist uncovers a notebook containing a list of names and numbers that no one has been able to decipher. & 0.017373 & 0.013657 & 0.026644 & 0.505205 \\
Salander, who is under state legal guardianship due to diagnosed mental incompetency, is appointed a new guardian, lawyer Nils Bjurman, after her previous guardian has a stroke. & 0.024011 & 0.011695 & 0.033138 & 0.461708 \\
Bjurman abuses his authority to extort sexual favors from Salander and then violently rapes her not realizing she has a hidden video camera in her bag. & 0.011282 & 0.011074 & 0.023850 & 0.462558 \\
At their next meeting she stuns him with a Taser, anally rapes him with a dildo and marks him as a rapist with a tattoo on his chest and stomach. & 0.017966 & 0.017839 & 0.027127 & 0.276713 \\ \midrule
Using her video recording she threatens blackmail, insisting that he write a glowing progress report and allow her full control of her money. & 0.014231 & 0.010573 & 0.000000 & -0.020144 \\
Blomkvist's daughter visits him and notes that the numbers from the notebook are Bible references. & 0.009230 & 0.006231 & 0.100033 & 0.099879 \\
\textbf{Blomkvist tells Vanger's lawyer Dirch Frode that he needs help with his research, and Frode recommends Salander based on the work she did researching Blomkvist himself.} & 0.024733 & 0.012166 & 0.072035 & 0.256393 \\
Blomkvist hires Salander to further investigate the notebook's content. & 0.009441 & 0.006414 & 0.051040 & 0.185852 \\
She uncovers a connection to a series of murders of young women that occurred from 1947 through 1967, with the women either being Jewish or having Biblical names; many of the Vangers are known antisemites. & 0.059293 & 0.016367 & 0.158320 & 0.070397 \\
During the investigation, Salander and Blomkvist become lovers. & 0.005047 & 0.005363 & 0.010820 & 0.164378 \\
Henrik's brother Harald identifies Martin, Harriet's brother and operational head of the Vanger empire, as a possible suspect. & 0.007916 & 0.014047 & 0.035330 & 0.142316 \\ \midrule
\textbf{Salander's research uncovers evidence that Martin and his deceased father, Gottfried, had committed the murders.} & 0.016496 & 0.022775 & 0.000000 & 0.230939 \\
\textbf{Blomkvist breaks into Martin's house to look for more clues, but Martin catches him and prepares to kill him.} & 0.026628 & 0.015342 & 0.095755 & 0.432990 \\
Martin brags about having killed women for decades, but denies killing Harriet. & 0.009883 & 0.005745 & 0.023151 & 0.110147 \\
Salander arrives, subdues Martin and saves Blomkvist. & 0.016412 & 0.009368 & 0.020565 & 0.308457 \\
While Salander tends to Blomkvist, Martin flees. & 0.004601 & 0.005421 & 0.007806 & 0.283848 \\
Salander, on her motorcycle, pursues Martin in his SUV. & 0.009073 & 0.014290 & 0.010464 & 0.109025 \\
He loses control of his vehicle on an icy road and dies when it catches fire. & 0.039593 & 0.020737 & 0.028716 & 0.102183 \\ \midrule
Salander nurses Blomkvist back to health, and tells him that she tried to kill her father when she was 12. & 0.011643 & 0.012107 & 0.000000 & 0.351047 \\
After recovering, Blomkvist deduces that Harriet is still alive and her cousin Anita likely knows where she is. & 0.014321 & 0.008425 & 0.048531 & 0.316454 \\
He and Salander monitor Anita, waiting for her to contact Harriet. & 0.009462 & 0.006949 & 0.017101 & 0.142604 \\
\textbf{When nothing happens, Blomkvist confronts her, correctly deducing that Anita is Harriet herself.} & 0.012247 & 0.008102 & 0.021557 & 0.273596 \\
She explains that her father and brother had sexually abused her for years, and that Martin saw her kill their father in self-defense. & 0.022848 & 0.011363 & 0.042934 & 0.079958 \\
Her cousin Anita smuggled her out of the island and let her live under her identity. & 0.017918 & 0.011077 & 0.021131 & 0.108173 \\
Finally free of her brother, she returns to Sweden and tearfully reunites with Henrik. & 0.010425 & 0.013125 & 0.016238 & 0.270244 \\ \midrule
As promised, Henrik gives Blomkvist the information on Wennerström, but it proves to be worthless. & 0.008507 & 0.007942 & 0.000000 & 0.408722 \\
Salander hacks into Wennerström's computer and presents Blomkvist with damning evidence of Wennerström's crimes. & 0.014928 & 0.013480 & 0.080743 & 0.502369 \\
Blomkvist publishes an article which ruins Wennerström, who flees the country. & 0.024830 & 0.012859 & 0.081431 & 0.527359 \\
Salander hacks into Wennerström's bank accounts and, travelling to Switzerland in disguise, transfers two billion euros to various accounts. & 0.029498 & 0.007885 & 0.037509 & 0.407171 \\
Wennerström is soon found murdered. & 0.009358 & 0.006529 & 0.016787 & 0.331802 \\
Salander reveals to her former guardian that she is in love with Blomkvist. & 0.003811 & 0.004793 & 0.006905 & 0.368140 \\
On her way to give Blomkvist a Christmas present, however, Salander sees him and his longtime lover and business partner Erika Berger walking together happily. & 0.013784 & 0.011499 & 0.012304 & 0.467294 \\
Heartbroken, she discards the gift and rides away. & 0.004822 & 0.004665 & 0.004822 & 0.233605 \\
\bottomrule
\end{longtable}
\end{small}
}

\twocolumn

\section{ROCStories Annotation Guidelines}
\label{sec:guidelines}

Based on learning lessons from \citet{ouyang-2015-modeling_mres} and \citet{ouyang-2017-crowd} shared with us, and feedback in a trial run, we first ask participants to write a short summary of the story. They are then instructed to use their summary to help determine the most important sentence.
When deciding between multiple possible sentences, participants were told to use the principle of selecting the sentence which is most amazing, astonishing or surprising.
The full guidelines are as follows:

\paragraph{1. Read the story}

\

\paragraph{2. Summarize the story}

\

In your summary, you should describe what is most memorable about the story.
Your summary should be a spoiler for the story.

\begin{itemize}
\itemsep0pt
\item It must be less than 12 words.
\item Use your own words -- do not copy large portions from the original story.
\item There is no need to introduce characters or settings in your summary, unless it is vital.
\item You should use the names of characters, where they are known. If the story is written in first person (`I', `My' etc.), write your summary in first person.
\end{itemize}

\paragraph{3. Identify the sentence which is most important to the story}

\

Look at your summary and decide which sentences include information used in your summary. 
Of these, select the sentence which is most \underline{amazing, astonishing or surprising}.

\begin{itemize}
\itemsep0pt
\item You may only choose one whole sentence.
\item If a sentence contains multiple parts, don't worry if part of the sentence is not memorable. 
\end{itemize}

\subsection{Example}

\ex.\label{ex:my-youngest}
\a. My youngest nephew came to stay with us for a weekend. \label{ex:my-youngest-1}
\b. He taught my youngest son lots of new words. 
\c. He showed him some next and interesting gestures. 
\d. \textbf{Finally, he gave us all a tutorial on tantrum throwing.} \label{ex:my-youngest-4}
\e. I was elated to see his parents come and pick him up.

Summary:

\begin{center}
\textbf{My nephew threw a tantrum over the weekend.}
\end{center}

The summary contains information from sentence~\ref{ex:my-youngest-1} and \ref{ex:my-youngest-4}. Sentence~\ref{ex:my-youngest-4} reveals that the `new words' and `interesting gestures' were part of tantrums he threw. This is the most surprising and astonishing sentence, so we select it.

\ex. \label{ex:ricky-was-g}
\a. Ricky was playing tennis with Bob. \label{ex:ricky-was-1-g}
\b. \textbf{Bob served the ball and hit Ricky right in the head!} \label{ex:ricky-was-2-g}
\c. Ricky fainted on the tennis court.
\d. Bob gave Ricky some water to drink.
\e. Ricky felt better after resting for awhile.

\begin{center}
\textbf{Bob hits Ricky in the head during a tennis match.}
\end{center}

The summary contains information from sentence~\ref{ex:ricky-was-1-g} and \ref{ex:ricky-was-2-g}. Sentence~\ref{ex:ricky-was-2-g} is the most amazingly bad and surprising sentence, so we select it.

\section{LLM Prompts}
\label{sec:llm_prompts}

\subsection{ROCStory Experiments}\label{sec:rocstory:prompts}

\citet{mostafazadeh-2016-roc_stories} gave crowd-workers the following prompt:

\begin{tcolorbox}[colback=white, colframe=gray!60!white, sharp corners, boxrule=0.5mm, title=ROCStories Human Prompt]
Imagine that you want to tell a five-sentence story to your friend. It can be about something that happened, something you or someone else has experienced in the past, or simply any life story about someone or something. Your task is to write this five-sentence story. Your story should have ALL of the following five properties.

\

1. Your story should be entirely realistic.

2. Your story should read like a coherent story, with a specific beginning and end, where something happens in between.

3. Each sentence in your story should be logically related to the next sentence and be about the characters of the story.
\end{tcolorbox}

We use the following prompt in order to generate a more verbose version of each story.

\begin{tcolorbox}[colback=white, colframe=gray!60!white, sharp corners, boxrule=0.5mm, title=LLM Prompt 1]
\{5-SENTENCE STORY\}

\

The above story was written in order to describe a real situation. Many things occurred that the author did not include in the story. Now imagine you know everything about the situation. Write an extended 20-sentence story about the situation. Your story should have ALL of the following five properties.

\

1. Your story should be faithful to the events that occurred in the original story.

2. You should re-write all original events in your own words.

3. All new events should be entirely realistic given the original story.

4. Each sentence in the original story should become four sentences in your extended story.

5. Every sentence should be simple and shorter than the sentences in the original story.

\

Respond now only with your extended story in prose.
\end{tcolorbox}

\ex. \label{ex:roc-example-long}
Ricky and Bob decided to play tennis together one sunny afternoon.\\
They were excited to enjoy the game.\\
Bob served the ball with great force.\\
Unfortunately, the ball hit Ricky squarely on the head.\\
Ricky was completely taken by surprise.\\
He stumbled back and felt dizzy.\\
The impact knocked the breath out of him.\\
He quickly fainted and fell to the ground.\\
Bob rushed over to Ricky immediately.\\
He was very worried about his friend.\\
Bob knelt beside him and checked for any serious injuries.\\
After a moment, he grabbed his water bottle.\\
He gently poured a little water into Ricky's mouth.\\
Ricky's eyes fluttered open slowly.\\
He took a few deep breaths to regain his composure.\\
The cool water helped him feel better.\\
They decided to take a break from playing.\\
Bob sat next to Ricky to keep him company.\\
They talked about what had happened and laughed a little.\\
Ricky teased Bob about his serve being too powerful.\\
After resting for a while, Ricky felt more like himself.\\
He stood up slowly and stretched his arms.\\
Bob encouraged him to take it easy.\\
With a smile, Ricky agreed to play again, but he would stand back a bit during serves.\\
They resumed their game, now more cautious than before.\\
Both were determined to have a good time despite the earlier mishap.\\
The day turned out to be a memorable one.\\
They left the court feeling happy and grateful for their friendship.

Story~\ref{ex:roc-example-long} shows the result for story~\ref{ex:roc-example} as input (page~\pageref{ex:roc-example}).

Based on these verbose stories, we prompt the same LLM in order to generate a new 5 sentence story, forming the narrative twin.
We use the following prompt:

\begin{tcolorbox}[colback=white, colframe=gray!60!white, sharp corners, boxrule=0.5mm, title=LLM Prompt 2]
\{20-SENTENCE STORY\}

\

Imagine that you want to tell the above story to your friend. Your task is to retell the story in five sentences. Your retelling should have ALL of the following five properties.

\

1. Your retelling should omit events that are not important to the provided story.

2. Your retelling should be faithful to the most important events occur in the provided story.

3. Your retelling should only include events that are in the provided story.

4. Every sentence should be logically related to the next sentence and be about the characters of the story.

5. Every sentence should be simple and short. Maximum one single clause.

\

Respond now only with your retelling in prose.
\end{tcolorbox}

Story~\ref{ex:roc-example-twin} shows the resulting twin for the example.

\ex. \label{ex:roc-example-twin}
\a. Ricky and Bob decided to play tennis one sunny afternoon.
\b. Bob served the ball, but it accidentally hit Ricky on the head, and he fainted.
\c. Bob quickly checked on Ricky and poured water into his mouth.
\d. Ricky woke up and felt better after resting.
\e. They continued to play, being more cautious this time.

For the negatives, we use the following prompt:

\begin{tcolorbox}[colback=white, colframe=gray!60!white, sharp corners, boxrule=0.5mm, title=LLM Prompt 3]
\{5-SENTENCE STORY\}

\

Your task is now to write a new version of the above five-sentence story. Your story should be very similar to the above story. But you should change the climax completely. Your story should have ALL of the following properties.

\

1. Your story should involve all of the same characters as the above story.

2. Your story should be based on the same setting as the above story.

3. The main and most important event in your new story should be fundamentally different from what occurred in the above story.

4. Other than the main and most important event, your new story should be very similar to the above story.

5. Every sentence in your new story should be logically related to the next sentence and be about the characters of the story.

6. Your story should be entirely realistic.

7. Your story should read like a coherent story, with a specific beginning and end, where something happens in between.

8. Every sentence in your new story should be simple and short. Maximum one single clause.

9. Your story should be exactly five sentences.

\

Respond now only with your story in prose.
\end{tcolorbox}

Story~\ref{ex:roc-example-negative} shows the resulting negative for the example.

\ex. \label{ex:roc-example-negative}
\a. Ricky was playing tennis with Bob.
\b. Bob served the ball and it landed just inside the baseline.
\c. Ricky ran quickly and returned the ball with a strong hit.
\d. They continued to play energetically for several more minutes.
\e. After the game, Bob gave Ricky some water to drink, and Ricky felt refreshed.

\newpage
\onecolumn
For our ROCStory annotator-style prompting, we use the following:

\begin{tcolorbox}[colback=white, colframe=gray!60!white, sharp corners, boxrule=0.5mm, title=ROCStories annotator-style prompt]
You will be given a short story. The story will always be 5 sentences long. You will be asked to write a short one-sentence summary of the story. You will then be asked to identify the sentence which is most important to the story.

\

In your summary, you should describe what is most memorable about the story. Your summary should be a spoiler for the story.

\

- It must be less than 12 words.

- Use your own words - do not copy large portions from the original story.

- There is no need to introduce characters or settings in your summary, unless it is vital.

- You should use the names of characters, where they are known. If the story is written in first person ('I', 'My' etc.), write your summary in first person.

\

For your identification of the sentence which is most important to the story, look at your summary and decide which sentences include information used in your summary. Of these, select the sentence which is most amazing, astonishing or surprising.

\

- You may only choose one whole sentence.

- If a sentence contains multiple parts, don't worry if part of the sentence is not memorable. 

\

Respond in the form of: {"summary": str, "most\_important": int}

\

1. Ricky was playing tennis with Bob.

2. Bob served the ball and hit Ricky right in the head!

3. Ricky fainted on the tennis court.

4. Bob gave Ricky some water to drink.

5. Ricky felt better after resting for awhile.

\

\{"summary": "Bob hits Ricky in the head during a tennis match.", "most\_important": 2\}

\

1. My youngest nephew came to stay with us for a weekend.

2. He taught my youngest son lots of new words.

3. He showed him some next and interesting gestures.

4. Finally, he gave us all a tutorial on tantrum throwing.

5. I was elated to see his parents come and pick him up.

\

\{"summary": "My nephew threw a tantrum over the weekend.", "most\_important": 4\}

\

1. {story sentence 1}

2. {story sentence 2}

3. {story sentence 3}

4. {story sentence 4}

5. {story sentence 5}

\end{tcolorbox}

\newpage
\twocolumn

For salience-style prompting, we use the following:

\begin{tcolorbox}[colback=white, colframe=gray!60!white, sharp corners, boxrule=0.5mm, title=LLM saliency prompt]
INSERT\_STORY

\

Your task is to determine how narratively salient each sentence in the above story is. Narrative salience is the likelihood that information in a sentence would be included by someone else telling the same story. It is a float greater than 0.0, less than 1.0, and unique for each sentence. Respond in the form

\

\{"sentence\_1": float, "sentence\_2": float, "sentence\_3": float, "sentence\_4": float, "sentence\_5": float\}
\end{tcolorbox}

We use a structured response in order to enforce that the LLM assigns a number between 0.0 and 1.0 for every sentence in the story.

\subsection{Wikipedia Narrative Experiments}\label{sec:wik:prompts}

The Wikipedia narratives are longer, so we minimally change the salience prompt as follows:

\begin{tcolorbox}[colback=white, colframe=gray!60!white, sharp corners, boxrule=0.5mm, title=LLM saliency prompt]
INSERT\_STORY

\

Your task is to determine how narratively salient each sentence in the above story is. Narrative salience is the likelihood that information in a sentence would be included by someone else telling the same story. It is a float greater than 0.0, less than 1.0, and unique for each sentence. Respond in the form

\

\{"sentence\_1": float, "sentence\_i": float, ..., "sentence\_NUM\_SENTS": float\}
\end{tcolorbox}

\newpage
\onecolumn

For our experiments with LLM-generated Wikipedia narrative negatives, we use the following prompt:

\begin{tcolorbox}[colback=white, colframe=gray!60!white, sharp corners, boxrule=0.5mm, title=Wikipedia Narratives Negative LLM Prompt]
INSERT\_STORY

\

Your task is now to write a new story that takes heavy inspiration from the above 5-paragraph story. The new story should change the most important parts of the above story. The least important parts should be very similar to the above story.

\

You must change ALL of the following.

\

1. Change the main opportunity that is presented to the characters (approximately 10\% of the way through the story; typically in section 1).

2. Change the main event that changes the plans and defines the goal (approximately 30\% of the way through story; typically in section 2).

3. Change the point of no return event (approximately 50\% of the way through; typically in section 3).

4. Change the major setback faced by the characters (approximately 70\% of the way through; typically in section 4).

5. Change the climax event (approximately 90\% of the way through; typically in section 5).

\

You must keep ALL of the following the same.

\

6. Keep all of the same characters as the above story.

7. Keep the same setting as the above story. Use the same names.

8. Keep all of the minor events the same. Make them a logical and realistic part of your new story.

\

Your story must have ALL of the following properties:

\

9. Your story should be approximately the same total length as the above story.

10. Your story should be exactly 5 sections. Section 1 of your new story should correspond to new version of section 1 in the above story; the same holds for sections 2-5.

11. Your story should be written in exactly the same style as the above story.

\

Respond now with your sections in prose.
\end{tcolorbox}

\end{document}